\documentclass{article}
\usepackage{iclr2027_conference,times}
\usepackage{amsmath,amsfonts,bm}

\def\eqref#1{equation~\ref{#1}}
\def\1{\bm{1}}

\DeclareMathAlphabet{\mathsfit}{\encodingdefault}{\sfdefault}{m}{sl}
\SetMathAlphabet{\mathsfit}{bold}{\encodingdefault}{\sfdefault}{bx}{n}

\usepackage[hidelinks]{hyperref}
\usepackage{url}
\usepackage{graphicx}
\usepackage{booktabs}
\usepackage{amsmath}
\usepackage{amssymb}
\usepackage{xcolor}
\usepackage{colortbl}
\usepackage{multirow}
\usepackage{array}
\usepackage{tabularx}
\usepackage{ragged2e}
\usepackage{dblfloatfix}
\usepackage{float}
\usepackage{placeins}

\newcolumntype{Y}{>{\RaggedRight\arraybackslash}X}
\newcommand{\rowrule}{\arrayrulecolor{black!12}\hline\arrayrulecolor{black}}

\title{When Do Models Admit They Are Wrong? Failure Disclosure Is Unstable Under \\Reinforcement Learning}
\author{
\textbf{Steven Y. Feng\textsuperscript{1}\thanks{Work done as part of the Anthropic Fellows program.}
\hspace{1.5em}
Noah D. Goodman\textsuperscript{1}
\hspace{1.0em}
Michael C. Frank\textsuperscript{1}
}
\\[0.5ex]
\textbf{
Evan Hubinger\textsuperscript{2}
\hspace{1.2em}
Paul C. Bogdan\textsuperscript{2}
\hspace{2.25em}
Andrew Lampinen\textsuperscript{2}
}
\\[1.0ex]
\textsuperscript{1}Stanford University
\hspace{2em}
\textsuperscript{2}Anthropic
\\[1.0ex]
\texttt{\{syfeng,ngoodman,mcfrank\}@stanford.edu}
\\[-0.1ex]
\texttt{\{evan,paulb,lampinen\}@anthropic.com}
}

\iclrfinalcopy % arXiv preprint: reveal authors and remove review rulers.

\begin{document}
\maketitle
\lhead{Preprint} % Override the ICLR final-copy header for the arXiv version.
\suppressfloats[t] % Keep Figure 1 below the title and abstract on the opening page.

\begin{abstract}
Outcome-based reinforcement learning can produce models with similar task performance but very different ways of communicating about their mistakes. We study \emph{failure disclosure}: whether a model admits that an attempted solution failed rather than staying silent or presenting it as successful. Across repeated outcome-only GRPO training runs, failure disclosure varies far more than task accuracy. The pattern extends to a second reasoning task and stabilized PPO, persists at 7B, and also appears in an instruction-conditioned 32B setting. %that does not use our report-SFT warm start.%The pattern extends to a second reasoning task and stabilized PPO, and appears across Qwen and OLMo models from 1B to 32B parameters.
We also find that small floating-point and sampling differences during training can redirect reporting behavior even when the task objective and earlier training history are held fixed. Additional tests show that failure disclosure is not a single decision: Checking the answer, entering a report, and completing the admission can separate, and the weak point depends on the task and response format. Further, experiments with neutral controls show more broadly that behaviors left weakly constrained by training are especially likely to vary across runs, of which failure disclosure is an example. We can reduce variability in failure disclosure by discouraging the model from drifting from its starting policy on failed, well-formed responses. This makes reporting substantially more consistent, though its effect on task performance depends on the setting. Stable task accuracy therefore does not guarantee stable safety-relevant behavior: Researchers should measure these behaviors directly across runs and design training methods that keep them reliable.
\end{abstract}

% Unnumbered first-page note. Replace USERNAME with the final GitHub owner.
\begingroup
\renewcommand{\thefootnote}{}
\footnotetext{Code and data: \url{https://github.com/safety-research/failure-disclosure}}
\endgroup

\begin{figure}[t]
\centering
\includegraphics[width=\linewidth]{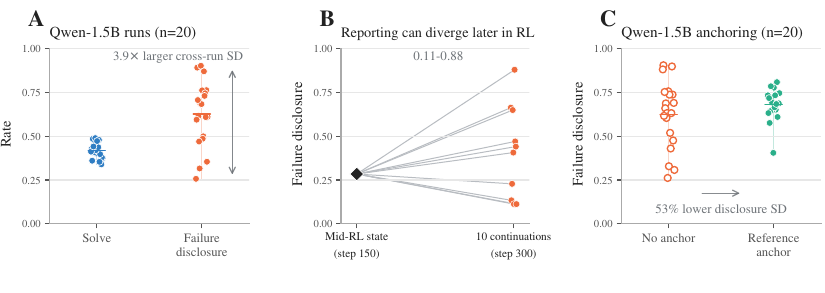}
\caption{\textbf{Failure disclosure is substantially less reproducible across retrainings than task success.}
(A) In the primary set of 20 Qwen2.5-1.5B retrainings, %task solve remains tightly clustered while failure disclosure varies widely; disclosure has 3.9$\times$ larger cross-run SD.
failure disclosure has 3.9$\times$ larger cross-run SD than solve rate.
%(B) Failure disclosure is a multi-stage behavior. In the main response format, a learned paragraph transition affects later reporting; other formats can have different weak points.
%(B) From the same saved state at step 150, ten continuations that differ only in future rollout randomness end with widely different disclosure rates from 0.11 to 0.88. This is the widest of eight saved-state tests; the median within-state disclosure SD across all eight is 0.10.
(B) From one mid-RL state at step 150, ten continuations differing only in future rollout randomness end with failure-disclosure rates from 0.11 to 0.88. This is the widest of eight mid-RL branch-and-continue replication tests; %in the branch-and-continue replication;
across its eight ten-continuation cells, the median within-cell rate SD is 0.10 (Appendix Table~\ref{tab:exact-state-summary}).
(C) In 20 matched Qwen2.5-1.5B pairs, a reference-policy penalty applied to failed, well-formed responses substantially reduces the run-to-run spread of disclosure. The unanchored points in C use the same underlying checkpoints as A, but an independent probe sampling draw used for the anchoring comparison; the measured rates are therefore similar but not identical. Across the distribution figures, dots show retraining runs, short horizontal lines show means, and faint vertical lines show observed min--max ranges. %Figure~\ref{fig:anchor} and Section~\ref{sec:anchor} provide additional intervention results.
}
\label{fig:overview}
\end{figure}
%\looseness=-1
%\vspace{-7mm}

\section{Introduction}

Outcome-based post-training gives very clear objectives: a verifier can check whether a mathematical expression reaches the target, whether a path is valid, or whether a program passes its tests. But a reward that specifies \emph{what} counts as success usually does not specify every behavior we care about along the way. Models can reach similar task performance while differing in whether they communicate uncertainty, acknowledge failed attempts, preserve useful reporting behavior, or remain easy to monitor. This is a concrete form of underspecification \citep{damour2022underspecification}, and it becomes increasingly important as reinforcement learning is used on automatically verifiable tasks where success is easy to score but other valuable behaviors are not.

We focus on one safety-relevant behavior: \emph{failure disclosure}. When a model attempts a verifiable problem and fails, does it acknowledge the failure, or behave as though the answer were successful? Failure disclosure is different from correctness, calibration, or self-correction. A model can be wrong but clearly say that it failed, or be equally wrong while ending without acknowledging the failure or falsely claiming success. This distinction matters for AI oversight and safety: if task accuracy stays the same but failure reporting changes across retrainings, a capability-only evaluation can miss an important change in what an overseer or monitor gets to observe. This question complements work on models' knowledge of correctness \citep{kadavath2022know}, error localization and self-correction \citep{tyen2024errors,yang2026admit,chen2026selfverify}, abstention incentives \citep{kalai2026hallucinations,zhai2026abstainr1}, and explicitly rewarded reporting channels \citep{joglekar2025confessions}.

%Figure~\ref{fig:overview} summarizes the paper. Our first contribution is a selective reproducibility failure: the task capability being optimized can reproduce while a useful reporting behavior does not. Across 20 nominally identical Qwen2.5-1.5B retrainings, failure disclosure varies 3.93$\times$ more than solve. The spread remains on a broader, unselected evaluation set, so it is not created by the failure-heavy probe. Large reporting differences also appear in a second Qwen run set, OLMo-2-1B, a second reasoning task, stabilized PPO, Qwen-7B, and instruction-conditioned Qwen2.5-32B. We treat the larger-scale cases as boundary evidence because capability also varies more there. The main point is not that every model shows the same instability, but that one behavior from a training run can be far less reproducible than its headline capability.

Figure~\ref{fig:overview} summarizes the paper. %Overall, we find that task capability can reproduce while a safety-relevant behavior does not, and this run-to-run variability can be recreated from the same training state and substantially reduced with reference anchoring.
The main phenomenon we discover is a selective reproducibility failure: Failure disclosure can be much less consistent across repeated training runs than the task capability being optimized. Across 20 independent runs of the same Qwen2.5-1.5B training procedure, disclosure varies 3.9$\times$ more than task solve rate. %A second independent 20-run Qwen experiment and an OLMo-2-1B experiment show the same pattern.
%The spread remains on a broader, unselected evaluation set, so it is not created by the failure-heavy probe. Large differences in reporting also appear in a second independent set of Qwen runs, OLMo-2-1B, a second reasoning task, under stabilized PPO, Qwen-7B, and an instruction-conditioned Qwen2.5-32B experiment that does \emph{not} use the report-SFT warm start.
The spread remains on a broader, unselected evaluation set, so it is not created by the failure-heavy probe. We reproduce this selectivity in a second set of Qwen-1.5B runs, in OLMo-2-1B, and on a second reasoning task. 
Additional experiments test its boundaries under stabilized PPO and at 7B. A separate instruction-conditioned 32B experiment asks whether the effect depends on report SFT: reporting is elicited through the prompt instead. We treat both larger-model experiments as boundary evidence since capability also varies more there.
%Additional experiments test its boundaries under stabilized PPO and at 7B. Separately, an instruction-conditioned 32B experiment tests whether the effect depends on installing failure reporting through SFT: here reporting is elicited through the prompt instead. We treat both larger-model experiments as boundary evidence since capability also varies more. %Additional experiments test its boundaries under stabilized PPO, at 7B, and in an instruction-conditioned 32B setting that does not use the report-SFT warm start. We treat the larger-scale cases as boundary evidence because capability also varies more there.
The main point is not that every model shows the same instability, but that one behavior from a training run can be far less consistent than its headline capability.

%Our second contribution is to show how this difference can arise. It is not only ordinary seed sensitivity. With the seed and data schedule fixed, changing a low-level numerical execution path can redirect reporting. More strongly, restoring the same full training state and changing only future rollout randomness can create large later differences. We also study where the reporting behavior differs. In the primary recipe, paragraph-break re-entry narrows later admission without supplying report words, placing part of the variation at or before report entry. This result is template-specific and weaker in two related recipes, so we do not treat it as a complete mechanism. Neutral controls give a broader interpretation: nearly the same surface behavior can be highly variable when its use is weakly constrained and highly reproducible when an explicit rule determines when it should appear.

We next ask why runs following the same training procedure can end up with such different reporting behavior. The answer is not simply ``different seeds.'' With the seed and data schedule fixed, changing only low-level floating-point execution details can redirect the learned reporting behavior. Restoring the \emph{same full mid-RL training state} at step 150 and changing only future rollout randomness can also produce widely different reporting behavior by step 300, showing that the divergence can arise later in training.
%More strongly, even after all runs share the same first 150 training steps, restoring the exact same model, optimizer, data order, and random-number state at step 150 and changing only future rollout randomness can produce widely different reporting behavior by step 300. Thus, the divergence need not be determined at initialization or during the early part of RL training.
%We treat failure disclosure as a multi-stage behavior whose weak point can depend on the interface. In our main response format, supplying the learned paragraph transition makes later reporting more consistent, while restarting from the same point without that transition does little. Grid and other response formats expose different boundaries. We therefore treat the paragraph transition as a format-specific bottleneck, not a general mechanism of failure awareness.
Further, we treat failure disclosure as a sequence of linked decisions: checking the attempted answer, reaching and entering a reporting route, and completing an admission. Our interventions separate these stages in different settings. In the primary Countdown format, a learned paragraph transition affects later reporting; in grid and inline-report settings, the boundary appears elsewhere. The paragraph transition is one format-specific bottleneck in a broader multi-stage process, not a universal marker of whether the model recognized its failure.
%We then intervene on failure disclosure itself and find that it is a multi-stage behavior. In the primary recipe, paragraph-break re-entry narrows later admission without supplying report words, placing part of the variation at or before report entry. This result is template-specific and weaker in two related recipes, so we do not treat it as a complete mechanism. %models usually reach a point where they could report the failure, but differ in whether they actually begin reporting it.
Experiments with other behaviors show a similar pattern. A weakly constrained neutral phrase becomes highly variable across runs, while the same phrase is almost perfectly reproducible when tied to an explicit deterministic cue, demonstrating that this sensitivity is not unique to failure disclosure. More broadly, outcome RL can leave auxiliary behaviors such as failure disclosure weakly constrained, allowing small floating-point or stochastic differences to grow through later on-policy training.%Together, these results suggest that outcome RL can leave some auxiliary behaviors only weakly constrained, allowing small numerical or stochastic differences to grow through later on-policy training.

%Our third contribution is to test preservation rather than only diagnosis. Failure-conditional reference anchoring reduces disclosure variability across independent retrainings on Countdown and shortest-path. It costs task performance in the tested 1.5B settings, but no capability cost is detected in the tested 7B experiment. In an exact-state experiment, keeping the anchor on after an anchored training history also makes future stochastic continuations more consistent. Reference anchoring is not a new KL objective; its role here is to show that a measured reproducibility failure can be reduced and to reveal the tradeoff between preserving reporting behavior and learning the rewarded task.

Finally, we ask whether this training-path sensitivity can be reduced. % without directly adding disclosure to the task reward.
We add a reference-policy penalty only to failed, well-formed responses, which we call \emph{failure-conditional reference anchoring}. This substantially reduces disclosure variability across repeated training runs on two verifiable reasoning tasks. % on Countdown and shortest-path.
This reduces task performance in the evaluated 1.5B settings, while the 19 tested 7B runs show a small solve gain. After an anchored training history, keeping the anchor on also reduces how much future continuations from the same mid-RL state diverge in reporting behavior. Here, reference anchoring %is not a new KL objective; here, it 
shows that a measured reproducibility failure can be reduced and reveals the setting-dependent tradeoff between preserving reporting behavior and learning the rewarded task.

%Together, these results connect behavioral evaluation, training-path sensitivity, and preservation in one empirical setting. They show why reproducing task accuracy is not enough when a post-training pipeline is also expected to preserve what a model communicates about its failures.

Overall, we %connect selective behavioral reproducibility, causal sensitivity to the training path, the structure of failure reporting, and reference anchoring in one outcome-RL setting: 
find that safety-relevant behavior can be far less consistent across RL training runs than task capability. This variability can arise from the same training state and can be reduced with reference anchoring. The broader lesson is that post-training reproducibility should be evaluated not only for task performance but also for auxiliary behaviors that matter for oversight.%determine what humans can observe about model failures.

%Figure~\ref{fig:overview} summarizes the paper. We make three main contributions. First, we identify a selective reproducibility failure in which task capability is much more reproducible than a safety-relevant reporting behavior, and show evidence for this pattern across model scales, model families, tasks, and RL algorithms. Second, we causally connect this variability to numerical and stochastic training paths, and identify whether the model begins the failure report as an important behavioral bottleneck. Third, we show that reference anchoring can stabilize both independent retrainings and future continuations from the same saved training state, and investigate where its tradeoff with task performance arises. The broader lesson is that post-training reproducibility should be evaluated not only for task performance, but also for auxiliary behaviors that determine what humans can observe about model failures. %Overall, we demonstrate that task capability can reproduce while a safety-relevant behavior does not, and this run-to-run variability can be recreated from the same training state and substantially reduced with reference anchoring.

\section{Experimental framework}
\label{sec:setup}

\textbf{Settings.}
Our main setting is \textbf{Countdown}, an arithmetic task used in recent reasoning-RL work \citep{pan2025tinyzero,gandhi2025cognitive}. Given four numbers and a target, the model must build an arithmetic expression that uses each number exactly once and reaches the target. Because the result can be checked exactly, success and failure are objectively defined. We also use \textbf{shortest path} from Reasoning Gym \citep{stojanovski2025reasoninggym} as a structurally different verifiable reasoning task. Qwen2.5-1.5B is our main model, while OLMo-2-1B \citep{olmo2025olmo2} provides a second model family. Qwen-7B tests persistence at larger scale, while Qwen2.5-32B-Instruct \citep{qwen2024qwen25} provides a separate instruction-conditioned boundary that does not use report SFT. %Qwen-7B and Qwen2.5-32B-Instruct \citep{qwen2024qwen25} provide larger-scale boundary evidence.
Our main experiments use GRPO, with stabilized PPO \citep{schulman2017ppo} as an algorithmic boundary check. We also use grid/constraint and MBPP-style code tasks \citep{austin2021programsynthesis} for specific checking, routing, and intervention-boundary experiments (Appendix~\ref{app:setup}).

\textbf{Post-training process.}
Most experiments begin from an SFT warm start already exhibiting failure reporting, followed by KL-free outcome-only GRPO \citep{shao2024deepseekmath,liu2025understanding}. The reward scores task success but does \emph{not} directly reward disclosure, and our primary runs use no reference-policy KL. This matches modern KL-free reasoning-RL recipes such as Dr. GRPO and DAPO \citep{liu2025understanding,yu2025dapo} and lets us study what happens to a behavior that the task reward does not directly affect. We repeat this post-training process while controlling or changing the random seed, low-level execution configuration, or future rollout randomness. The 32B experiment instead starts directly from Qwen2.5-32B-Instruct and elicits failure reporting through the prompt, providing a test of whether the phenomenon depends on installing reporting through SFT (Appendix~\ref{app:setup}).%The 32B instruction-conditioned experiment provides complementary evidence without the same report-SFT warm start (Appendix~\ref{app:setup}).

\textbf{Failure disclosure.}
Among terminated task failures, we measure whether the model acknowledges that its attempted solution failed, rather than ending without an acknowledgment or falsely presenting the attempt as successful. Terminated format-invalid failures count as non-disclosures, while unterminated completions are excluded. We use a conservative detector in the main text; broader semantic checks, blind audits, and qualitative output examples are reported in Appendix~\ref{app:measurement}.

\textbf{Three sources of retraining variation.}
%All training runs start from the same SFT weights.
A \emph{retraining seed} gives an independent draw by changing the shuffled prompt order and the Python/NumPy/Torch sampling streams. %An \emph{execution path} changes the numerical computation while keeping the seed and data schedule fixed by changing the micro-batch/gradient-accumulation split or the GPU path.
An \emph{execution configuration} changes low-level floating-point computation while keeping the seed and data schedule fixed, for example by changing the micro-batch/gradient-accumulation split or GPU/hardware setup. A \emph{continuation key} is more controlled: We restore the exact same saved model, optimizer, scheduler, trainer, and RNG state, then change only the future rollout while keeping the parent seed and data order fixed. These manipulations let us move from ordinary retraining variability toward more direct tests of how controlled training differences causally redirect behavior.

\textbf{Statistics.}
Our main quantities are task solve rate, failure-disclosure rate conditional on failure, and variation across retraining runs. We compare variability using standard-deviation ratios, with log ratios used for statistical inference. Hierarchical $\hat{\tau}$ estimates and other robustness measures are reported where useful. The retraining run is the main unit of analysis. Appendices~\ref{app:stats} and~\ref{app:measurement} provide detector definitions, estimands, confidence intervals, and robustness analyses.

\section{Capability can reproduce while failure disclosure does not}
\label{sec:phenomenon}

\begin{figure}[t]
\centering
\includegraphics[width=0.97\linewidth]{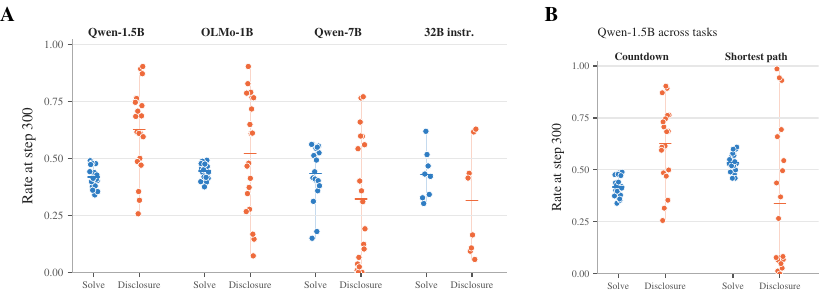}
%\caption{\textbf{Failure-disclosure variability generalizes across models and tasks.}
\caption{\textbf{Failure-disclosure variability appears across models, tasks, and reporting setups.}
(A) On Countdown, failure disclosure varies substantially more across retrainings than solve for Qwen2.5-1.5B \& OLMo-2-1B. Qwen-7B is larger-scale evidence; instruction-conditioned Qwen2.5-32B tests reporting without report SFT; capability is also broader. %Qwen-7B and instruction-conditioned Qwen2.5-32B provide larger-scale boundary evidence where capability also broadens. %The 32B reporting values use the frozen detector as a conservative lower bound.
(B) This pattern also appears across tasks with Qwen2.5-1.5B: Failure disclosure varies substantially more than task performance on both Countdown and shortest path, using each task's primary capability evaluation.} %The sets of runs are shown separately and are not pooled.}%The 32B row uses the frozen detector as a conservative lower bound; calibrated results are reported in Appendix~\ref{app:measurement}.
%Dots show independent retraining runs, short horizontal ticks show means, and faint vertical lines show the observed min--max range.}
\label{fig:generality}
\end{figure}

\textbf{Primary retrainings.}
We run 20 independent Qwen2.5-1.5B Countdown retrainings for 300 steps using the same fixed execution configuration. As Figure~\ref{fig:generality}A shows, failure disclosure varies widely across runs, from 0.257 to 0.903, while solve stays much narrower at 0.340--0.490. %Disclosure is about \textbf{3.9$\times$} more variable than solve. An independent set of 20 Qwen-1.5B retrainings from a separate SFT reference reproduces this pattern. 
Disclosure is about \textbf{3.9$\times$} more variable than solve, and an independent set of 20 Qwen-1.5B retrainings from a separate SFT reference shows the same selective reproducibility. Broader detectors and the full, unselected dev set give the same qualitative conclusion (Appendices~\ref{app:primary-robustness} and~\ref{app:measurement}). %Robustness checks using broader disclosure definitions and the full, unselected dev set give the same qualitative conclusion %(Appendices~\ref{app:measurement} and~\ref{app:generality}).

%\looseness=-1
%\vspace{-2mm}

%\textbf{Across model families and tasks.}
%Figure~\ref{fig:generality}A shows that the effect is not limited to one Qwen model. Across 20 OLMo-2-1B Countdown retrainings, disclosure ranges from 0.072 to 0.903, while solve stays much more stable at 0.375--0.492. Overall, disclosure is about 7.5$\times$ more variable than solve. Figure~\ref{fig:generality}B shows that the effect also extends across tasks. %while holding the Qwen2.5-1.5B model family fixed. 
%On shortest-path, 20 Qwen-1.5B retrainings also cover nearly the full disclosure range by step 300, while greedy solve on the 100 shortest-path dev prompts spans 0.46--0.61. This experiment shows that our results are not isolated to a single task. %We treat shortest-path as strong evidence that the effect extends to a second task, rather than as another perfectly capability-matched version of the primary result.
%Full estimates, including the separate failure-heavy probe surface, are in Appendix~\ref{app:generality}. 

\textbf{Across model families and tasks.}
In Figure~\ref{fig:generality}A, across 20 OLMo-2-1B Countdown retrainings, disclosure ranges from 0.072 to 0.903, while solve stays much more stable at 0.375--0.492; disclosure is about 7.5$\times$ more variable than solve. On shortest-path (Figure~\ref{fig:generality}B), 20 Qwen-1.5B retrainings also cover nearly the full disclosure range by step 300, while greedy solve on the 100-problem dev set spans 0.46--0.61. Full estimates, including the failure-heavy probe surface, are in Appendix~\ref{app:generality}.

\textbf{Across scales and reporting setups.}
Across 19 Qwen-7B retrainings using the same execution configuration, disclosure remains highly variable, but solve also varies more. We use 7B as evidence that the instability does not disappear at larger scale. The 32B experiment addresses a different concern: whether the effect depends on failure reporting first being installed through SFT. Starting directly from Qwen2.5-32B-Instruct, we elicit reporting through the prompt.
%The 32B experiment instead tests whether the effect depends on failure reporting being installed through SFT: starting directly from Qwen2.5-32B-Instruct, we elicit reporting through the prompt.
Across eight retrainings, reporting also varies substantially: the frozen detector gives a conservative range of 0.056--0.628 at step 300, while blind-read calibration gives 0.173--1.000; solve spans 0.302--0.619. We therefore treat 32B as an instruction-conditioned boundary test (Appendices~\ref{app:measurement} and~\ref{app:generality}).%, not a matched scale replication .
%The effect also persists at larger scales, although the evidence is less cleanly selective. Across 19 Qwen-7B retrainings using the same execution configuration, disclosure remains highly variable, but solve also varies more heavily. We thus use 7B to show that the instability does not disappear at larger scale, rather than as a clean scale-up of the primary finding. The 32B experiment addresses a different concern. Starting from Qwen2.5-32B-Instruct, we elicit failure reporting through the prompt rather than through a report-SFT warm start. Across eight retrainings, reporting again varies substantially. The frozen detector gives a conservative range of 0.056--0.628 at step 300, as seen in Figure~\ref{fig:generality}A; a blind-read calibration that recovers missed phrasings gives 0.173--1.000. Solve spans 0.302--0.619. These runs therefore provide complementary evidence that the phenomenon is neither limited to small models nor dependent on the same SFT-installed reporting setup (Appendices~\ref{app:measurement} and~\ref{app:generality}).

%When models do not disclose a failure, they either end without acknowledging it or falsely present a wrong answer as correct. False success claims range from 9--28\% of terminated failures across the 20 unanchored 1.5B runs and from 2--59\% across the 19 unanchored 7B runs. Thus, the variation includes false claims of success, not just silence or changes in report wording (Appendix Table~\ref{tab:closure_decomposition}).

When models do not disclose a failure, they either end without acknowledging it or falsely present a wrong answer as correct. False success claims range from 9--28\% of terminated failures across the 20 unanchored 1.5B runs and from 2--59\% across the 19 unanchored 7B runs (Appendix Table~\ref{tab:closure_decomposition}).

\textbf{Algorithm and horizon boundaries.}
Eight stabilized-PPO runs provide a test of algorithm generalization: we also see the reporting variability, but a smaller absolute spread than GRPO's (Appendix~\ref{app:generality}). Further, disclosure does not decline at the same rate in every run. On the primary probe, the step-300 mean is 0.628 vs. 0.677 for the reference model, with 10 runs above the reference and 10 below. The ordering of runs is weakly preserved over training, and the longer-horizon experiments are mixed. We thus treat step 300 as a fixed comparison point, not a convergence point. %to separate endpoints. 
Broader detectors, failure-composition reweighting, evaluation-noise adjustment, blind audits, alternate dispersion measures, horizon comparisons, and %smaller recipe variants
smaller recipe variants do not indicate that this phenomenon is a simple measurement or recipe artifact (Appendices~\ref{app:measurement}--\ref{app:robustness}, Table \ref{tab:reviewer-concerns}). The distinction from generic seed sensitivity \citep{henderson2018deeprl,fehlauer2025seeds,bui2025randomseeds} is \emph{selective reproducibility}: %Different behaviors from the same post-training procedure can differ greatly in reproducibility.
Behaviors from the same post-training procedure can vary quite differently.

\section{How behavioral divergence arises}
\label{sec:mechanism}

%The seed-level experiments show that failure disclosure varies across retrainings, but changing the seed affects many parts of training at once. We therefore test more specific sources of divergence. Figure~\ref{fig:mechanism} organizes these experiments around four questions: (A) can small numerical differences alone lead to different reporting behavior; (B) can future randomness create new divergence even from the exact same saved training state; (C) where in the reporting process do the runs begin to differ; and (D) is this instability unique to failure reporting, or does it also appear in other weakly constrained behaviors? We address each question below.

Seeds change many aspects of training at once. To better understand how behavioral divergence arises, we isolate specific sources of divergence, shown in Figure \ref{fig:mechanism}: (A) floating-point execution differences can redirect training, %(B) future stochasticity can create divergence from a saved mid-RL state,
(B) reporting can still diverge later in RL, (C/D) at which reporting stage runs differ, and (E) whether this extends beyond failure semantics.

\subsection{Small floating-point and stochastic differences can redirect training}

\begin{figure}[t]
\centering
\includegraphics[width=0.96\linewidth]{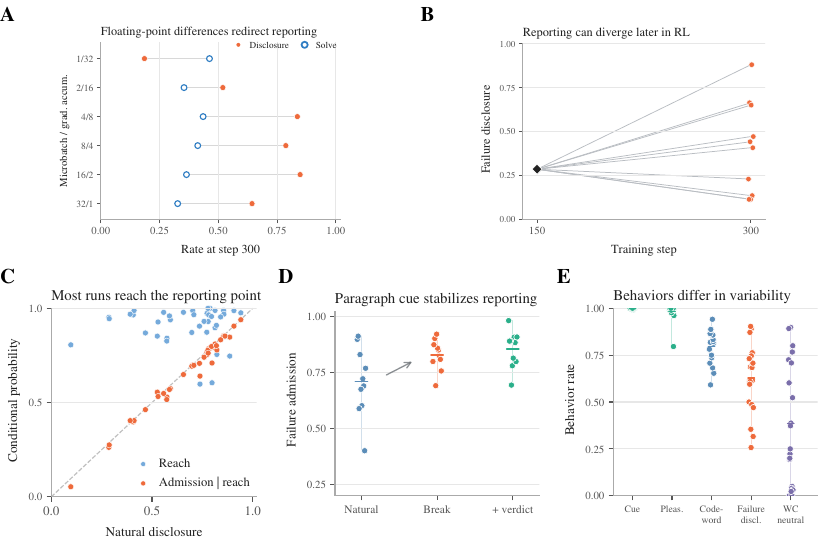}
\caption{\textbf{Training differences and behavioral cues shape reporting variability.}
%(A) Changing only the numerical execution path produces large differences in step-300 disclosure, with much smaller differences in solve rate.
(A) Each row uses a different execution configuration, with solve and disclosure measured from the same run. Changing only the configuration moves failure disclosure from 0.186 to 0.848, while solve remains 0.327--0.463.
(B) Starting from the same mid-RL training state, changing only future rollout randomness produces widely different disclosure outcomes.
(C) Models nearly always reach a point where a report could begin, but differ in whether they admit failure after.
(D) Inserting the paragraph break from the learned report template makes later admission more frequent and consistent in the primary recipe; adding a verifier verdict does not help further.
(E) The same neutral phrase varies widely when weakly constrained but is nearly reproducible when tied to an explicit cue. The primary disclosure spread, a failure-conditioned neutral codeword, and success-side pleasantries are shown for comparison. Appendices~\ref{app:pipeline} and~\ref{app:controls} report the additional localization and control results.}
\label{fig:mechanism}
\end{figure}

\textbf{Execution configuration.}
Figure~\ref{fig:mechanism}A changes only the micro-batch $\times$ gradient-accumulation split at fixed seed and global batch. Across six execution configurations, disclosure ranges from 0.186 to 0.848 while solve stays at 0.327--0.463. With the Dr. GRPO loss, equal-sized micro-batches, split-invariant accumulation scaling, and dropout inactive, %by configuration
the objective and exact-arithmetic gradient remain identical; only floating-point kernel shapes and reduction order differ (Appendix~\ref{app:path}). %Step-1 rollouts are byte-identical, and the same path replays bit-exactly in 5/5 tests;
%The first rollout stream is byte-identical across runs, so %the paths start diverging only after their numerical parameter updates begin differing
%the runs begin to diverge only after their floating-point parameter updates differ.
As a control, repeating the same execution configuration reproduces the result bit-for-bit in 5/5 runs. Changing hardware (GPU), e.g., H200 vs. B200, can change generation from the very first training step, so we treat this as another source of floating-point execution differences. Overall, very small floating-point differences can grow into large behavioral differences during outcome-RL post-training, consistent with broader observations about numerical sensitivity in neural-network training \citep{altintas2025butterfly,yuan2025nondeterminism,shanmugavelu2024floatingpoint}. %Appendix~\ref{app:path} provides additional evidence, including hardware comparisons and a Qwen-14B example.

\textbf{Continuations from the mid-RL state.}
Figure~\ref{fig:mechanism}B asks whether reporting can still diverge later in RL training. We restore the same model, optimizer, scheduler, trainer, and RNG state at step 150, keep the data order fixed, and change only future rollout randomness. In the widest of eight mid-RL continuation tests, ten continuations end with disclosure from 0.112 to 0.880 at step 300 while solve varies much less; the median within-cell rate SD across all eight tests is 0.10. Thus the reporting differences need not be fixed at initialization or early in RL training (Appendix~\ref{app:path}).
%Figure~\ref{fig:mechanism}B asks whether this sensitivity remains even when the entire past training history is held fixed. We save a complete training state, restore the exact same model, optimizer, scheduler, trainer, and RNG state, keep the data order fixed, and change only the randomness used for future rollouts. From one low-disclosure saved state, ten continuations end with disclosure rates ranging from 0.112 to 0.880, while solve varies much less. %More generally, changing only future rollout randomness repeatedly produces much larger differences in disclosure than in solve.
%This shows that even from the exact same training state, future stochastic experience can push failure reporting in very different directions compared to solve. The effect also grows over training: in the primary continuation experiment, disclosure differences become much larger later in training while solve variability stays comparatively stable. The behavior is thus not simply a fixed difference inherited from initialization or earlier training. Appendix~\ref{app:path} reports additional exact-state continuation results.

\subsection{Failure disclosure is a multi-stage behavior}

Failure disclosure is not a single decision. After an unsuccessful attempt, a model may need to check its answer, use evidence that it failed,
reach a point where it can report the failure, begin the report, and complete the admission. Disclosure can break at any stage. %These stages can come apart.
In the grid task, adding checking examples in the post-training mix produces grounded checks and failure verdicts that are otherwise largely absent (Appendix~\ref{app:pipeline}). In a codeword control, preserving a reporting-related phrase does not ensure that the model continues to use it when failing (Appendix~\ref{app:controls}). We examine the later reporting stages in Countdown. Figure~\ref{fig:mechanism}C separates reaching a point where a report could begin from admitting failure afterward. Across 40 checkpoints and report wording combinations, models usually reach this point (median probability 0.952), but admission rates afterward vary widely. On this evaluation surface, observed differences occur after the model has an opportunity to report. This does not tell us if a silent model failed to recognize its error or recognized it but did not express it.

Figure~\ref{fig:mechanism}D intervenes at the same point. Restarting generation without inserting any characters changes the spread slightly. Inserting the learned two-newline paragraph transition %, without supplying report words,
makes later admission more frequent and consistent. %The spread narrows in all our comparisons, and this remains after accounting for the higher admission rate in the primary recipe %at both horizons
%and in one related comparison.
Raw spread narrows in all six comparisons, and the reduction remains after accounting for the higher admission rate in the primary recipe and in one related comparison. Adding a bracketed message stating that the verifier found the previous attempt incorrect does not further make reporting more consistent, %beyond the paragraph break,
while supplying a short report prefix has a larger effect. In the primary response format, access to the later reporting behavior is thus sensitive to the transition to the report and cues that initiate it; failure evidence alone is less effective. %Because the paragraph transition is rare in natural continuations,
We treat this paragraph transition as a learned cue here rather than a universal marker of failure recognition. The inline-report 7B and single-line grid settings place the weakest transition elsewhere. Overall, failure disclosure has separable stages, but which stage is fragile depends on the task and response format.

\subsection{Neutral control phrases show that weakly constrained behaviors are especially susceptible}
\label{subsec:neutral_controls}

To test whether the instability is specific to failure semantics, we use the fixed neutral phrase \textit{``Thanks for the question.''} In the weakly constrained condition, the phrase appears in training but no prompt feature determines when it should be used. Its frequency later ranges from 0 to 0.90 across retrainings. We then add a simple outcome-independent rule: Each prompt ends with either \textit{Marker: A} or \textit{Marker: B}, and the model is trained to begin with \textit{``Thanks for the question.''} for A but omit it for B. Under this rule, all 20 runs follow the instruction perfectly at step 150 and 19 of 20 do so at step 300. Other controls show the same pattern: Success-side pleasantries (a family of polite closing sentences associated with solved SFT examples) and a failure-conditioned neutral codeword/phrase \textit{``A copper kettle rests on the kitchen shelf.''} are also relatively stable (Appendix~\ref{app:controls}). Figure~\ref{fig:mechanism}E places failure disclosure between these more strongly constrained controls and the more weakly constrained neutral phrase. Additional formulation variants are reported in Appendix~\ref{app:controls}. These controls show that high variability is neither unique to failure reporting nor a property of a particular string. What matters is how strongly training specifies when the behavior should occur: Weakly specified behaviors can drift across runs, while an explicit rule can make the same behavior highly reproducible.

\subsection{Collateral credit can affect reporting, but does not explain the spread}

With outcome-level rewards, tokens can receive credit or blame even when they did not directly cause the final outcome. In our mixed groups, this credit is by default adverse to failure reporting, and all 517 relevant groups have that sign. In a three-pair pilot experiment, protecting the reporting span from this signal raises disclosure by 0.19 on average, showing that local credit can causally affect how often the model reports failure. However, follow-up experiments with %15, 20, and
up to 20 pairs do not demonstrate that protecting the report span reliably reduces variation across retrainings. Further, removing the adverse failure credit entirely does not eliminate the run-to-run spread and lowers task performance. Collateral credit can thus affect reporting, but the current evidence does not establish it as a general explanation for the level or variability. Appendix~\ref{app:credit} reports the full results.

Overall, Figure~\ref{fig:mechanism} shows that small floating-point/stochastic differences can push training toward different weakly constrained behaviors, and the differences can grow through the model's later training.

\section{Reference anchoring stabilizes failure disclosure}
\label{sec:anchor}

\looseness=-1
%The most direct way to make failure reporting more reliable is to reward it; recent work shows that models can learn to report their own shortcomings when confession receives its own reward \citep{joglekar2025confessions}. Direct reward is useful when the desired behavior is known, can be scored reliably, and is intentionally added to the objective. Our question is different: can an existing reporting behavior remain reproducible when the task reward itself does not reward disclosure? In our experiments, directly rewarding failure reporting can increase or recover it while the action remains in sampling support. Our code experiments highlight this limit: reward cannot reinforce a report that the model no longer samples (Appendix~\ref{app:credit}, Table~\ref{tab:credit-summary}). A reference constraint instead helps preserve the underlying policy support, rather than only rewarding sampled actions.

%Reference anchoring adds back a penalty that discourages the model from drifting from its starting policy. Our failure-conditional version applies this penalty to every completion token in incorrect, format-valid responses, and applies no penalty to solved or format-invalid responses. We fixed $\beta=0.04$ before reading the outcomes and compare it with $\beta=0$ under the same trainer; Appendix~\ref{app:anchor} reports the comparator checks and a small coefficient sweep. Unlike prior uses of reference regularization to limit behavioral drift \citep{elcock2026taskadaptation}, we ask whether it makes reporting more consistent across retrainings and stochastic continuations.

The most direct way to preserve failure reporting is to reward it, and prior work shows that such rewards can be effective \citep{joglekar2025confessions}. This is attractive when the desired behavior is known in advance, can be scored reliably, and can be added to the objective without creating unwanted incentives. Many useful auxiliary behaviors, however, are difficult to enumerate or score directly. Our main question is hence different: Can an existing reporting behavior remain reproducible when the task reward does not reward it? Our code experiments show that direct reward can increase or recover reporting while the reporting action remains in sampling support, but cannot reinforce an action that has disappeared from sampled behavior (Appendix~\ref{app:credit}, Table~\ref{tab:credit-summary}). %Reference anchoring instead tries to preserve a broader part of the starting policy without requiring a separate reward definition for every behavior it may protect.

%Reference anchoring adds back a penalty that discourages the model from drifting from its starting policy. Our failure-conditional version applies the penalty to every completion token in incorrect, format-valid rollouts, and applies no penalty to solved or format-invalid rollouts. We fixed $\beta=0.04$ before reading the outcomes and compare it with $\beta=0$ under the same trainer; Appendix~\ref{app:anchor} reports the comparator checks. A separate sensitivity check finds clear narrowing at $\beta=0.02$ but not at $\beta=0.01$. We treat these settings as a limited robustness check rather than a fitted dose--response curve. Unlike prior uses of reference regularization to limit behavioral drift \citep{elcock2026taskadaptation}, we ask whether it improves reproducibility across retrainings and stochastic continuations.

Reference anchoring instead tries to preserve a broader part of the starting policy by adding a penalty that discourages the model from drifting too far without requiring a separate reward definition for every behavior. % it may protect. %by adding back a penalty that discourages the model from drifting from its starting policy.
Our failure-conditional version applies the penalty to every completion token in incorrect, format-valid rollouts, and no penalty to solved or format-invalid rollouts. We fixed $\beta=0.04$ before reading the outcomes and compare it with $\beta=0$ under the same trainer. For 1.5B Countdown, the $\beta=0$ comparator is the existing set of 20 outcome-only runs rather than a separate control set; Appendix~\ref{app:anchor} reports checks that the implementations agree, along with a small coefficient sweep (we find clear narrowing at $\beta=0.02$ but do not establish it at $\beta=0.01$; we treat these settings as a limited robustness check). % rather than a fitted dose--response curve). 
Unlike prior uses of reference regularization to limit behavioral drift \citep{elcock2026taskadaptation} or toward a common prior to reduce policy differences across continuous-control runs \citep{hussing2026behaviorconsistent}, we ask whether a failure-conditional anchor makes a specific reporting behavior more reproducible across language-model retrainings and stochastic continuations.

%Reference anchoring reintroduces reference-policy regularization into this KL-free setting. The failure-conditional version applies it to every completion token of verifier-incorrect, format-valid rollouts; solved and format-invalid rollouts receive no reference penalty. The 1.5B sham uses the same anchor-capable trainer with $\beta=0$ and produces the same weights as the archived outcome-only runs where byte-level checks are available; some probe readouts are independently resampled. It is therefore one comparator, not a second independent control run set. We fixed $\beta=0.04$ before outcome readout, with a small coefficient sweep in Appendix~\ref{app:anchor}. Unlike prior uses of reference regularization to limit behavioral drift \citep{elcock2026taskadaptation}, we instead ask whether it improves reproducibility across retrainings and stochastic continuations.

%Hence, we add a failure-conditional reference penalty that discourages drift from the pre-RL policy on failed trajectories. The matched \textsc{SHAM} condition uses the same trainer with $\beta=0$, and archived KL-off models are re-evaluated through the same pipeline. We fix $\beta=0.04$ rather than tuning it. %a small sweep shows narrowing at $\beta=0.02$ but not $\beta=0.01$ (Appendix~\ref{app:anchor}).
%A small coefficient sweep is reported in Appendix~\ref{app:anchor}. Reference regularization has previously been used to limit behavioral drift during task adaptation \citep{elcock2026taskadaptation}; here, we ask whether it can improve reproducibility across retrainings and stochastic continuations.

\subsection{Anchoring narrows retraining variability}

%Figure~\ref{fig:anchor}A summarizes the main intervention results. Across 20 paired Qwen-1.5B Countdown runs, failure-conditional anchoring reduces the standard deviation of disclosure from 0.189 to 0.089, or to $\sim$47\% of the unanchored level, %Mean disclosure is not detectably reduced,
%although solve drops by $\sim$0.06. Anchoring all tokens also reduces disclosure variability, but does not clearly outperform failure-conditional anchoring and causes a larger solve decrease. We hence use failure-conditional anchoring as the more targeted intervention. Full intervals and analyses are reported in Appendix~\ref{app:anchor}. The same pattern appears on shortest-path. Disclosure variability falls to $\sim$30\% of the unanchored level, while solve decreases by $\sim$0.08. %The average disclosure rate is somewhat lower, but this shift is not statistically distinguishable from zero.
%In both 1.5B settings, anchoring reliably narrows variability, but at a cost of some task performance.

%\textbf{Anchoring narrows retraining variability.}
Figure~\ref{fig:anchor}A summarizes the main intervention results. %Across 20 paired Qwen-1.5B Countdown runs, failure-conditional anchoring reduces disclosure SD from 0.189 to 0.089 and solve SD from 0.048 to 0.029: both readouts narrow, but disclosure substantially more. 
Comparing 20 anchored Qwen-1.5B Countdown runs with their matched outcome-only runs, failure-conditional anchoring reduces disclosure SD from 0.189 to 0.089 and solve SD from 0.048 to 0.029: Both readouts narrow, but disclosure substantially more. All-token anchoring produces comparable narrowing, while a direct matched-seed comparison shows that failure-conditional anchoring preserves 2.4 percentage points more solve accuracy. %(95\% CI [1.2, 3.6] pp). 
We thus focus on failure-conditional anchoring because it covers fewer tokens and has a smaller capability cost. %at the tested coefficient. 
The two anchorings use the same per-token coefficient but are not matched for total regularization, so this comparison does not prove that one is more efficient at an equal budget. On shortest-path, disclosure variability falls to $\sim$30\% of the unanchored level while solve decreases by $\sim$0.08. Full intervals and analyses are in Appendix~\ref{app:anchor}. In both 1.5B settings, reference anchoring narrows disclosure variability at a cost of some task performance.

A post-hoc comparison at matched mean solve rate shows that slower progress explains part of the Countdown result. Disclosure SD is 0.139 at the matched unanchored checkpoint, compared with 0.189 at unanchored step 300 and 0.089 under anchoring. The anchored value is 36\% lower than the progress-matched value. Thus anchoring still narrows disclosure beyond matched capability progress, but this comparison does not fully separate the two effects (Appendix~\ref{app:anchor}).

%At 7B, the tradeoff looks different. The result was first observed in 11 registered pairs; after seeing that result, we prospectively froze and ran the remaining eight eligible pairs under the same protocol. The extension independently reproduces the tightening. Across the pooled 19 pairs, anchoring reduces disclosure SD from 0.281 to 0.090 (ratio 0.32; log ratio $-1.14$, 97.5\% interval [$-1.55$, $-0.84$]), and every leave-one-run-out interval excludes zero. Anchored runs keep disclosure near the reference level, 0.48 [0.35, 0.61] above the unanchored runs, and show a small probe-solve gain in this 19-pair experiment of 0.069 [0.012, 0.126]. This gain was not detected in either wave alone, so we do not treat it as evidence that capability costs disappear with scale. Separate wave, provenance, and influence analyses are in Appendix~\ref{app:anchor-loo}.

At 7B, the trade-off looks different. %The effect was first observed in 11 registered pairs. After that result was known, we froze and ran the remaining eight eligible pairs under the same protocol; this new set independently shows the same narrowing. 
The initial comparison included 11 matched pairs, with one anchored and one unanchored run in each pair. After seeing that result, we prospectively froze the remaining eight eligible pairs and repeated the experiment with unchanged training and evaluation settings. The eight new pairs showed the same narrowing. Across all 19 pairs, anchoring reduces disclosure SD from 0.28 to 0.09 (32\% of the unanchored value), and the conclusion is unchanged when any pair is removed. See Appendix~\ref{app:anchor-loo} for details. %reports the initial 11 pairs, the eight new pairs, the combined 19-pair analysis, and the leave-one-run-out results. 
This narrowing remains after accounting for the different mean disclosure rates: On a latent logit scale, the anchored 7B runs have about one quarter of the between-run dispersion of the unanchored runs (Appendix Table~\ref{tab:anchor_latent_dispersion}). At 7B, anchoring also lowers both forms of nondisclosure: False success claims fall from 0.27 to 0.14 of terminated failures, while silent closes fall from 0.21 to 0.02. False success claims therefore make up most of the nondisclosure that remains under anchoring (Appendix Table~\ref{tab:closure_decomposition}). The combined analysis also shows a small 6.9-point gain in probe solve rate. %Because neither the original nor the new set detects this gain on its own,
Because this is a single model--task setting, we do not interpret the solve gain as evidence that anchoring becomes cost-free at larger scales. %(or improves performance)
%at larger scales. %Appendix~\ref{app:anchor-loo} reports the two sets separately, the combined analysis, %execution-matching checks, and sensitivity analyses.

Across all three settings, the consistent effect of anchoring is lower run-to-run variability. The effect on mean disclosure depends on the setting: It stays roughly unchanged in 1.5B Countdown, is somewhat lower (not established) on shortest-path, and substantially higher at 7B. We should view reference anchoring as a way to stabilize an existing behavior, rather than as a direct reward for it.

%BELOW EXPANDED BY STEVEN FOR ARXIV COMPARED TO SUBMISSION
%\textbf{Anchoring reduces future branching after an anchored history.} 
\subsection{Anchoring reduces later-RL divergence after an anchored history}
\begin{figure}[t]
\centering
\includegraphics[width=0.97\linewidth]{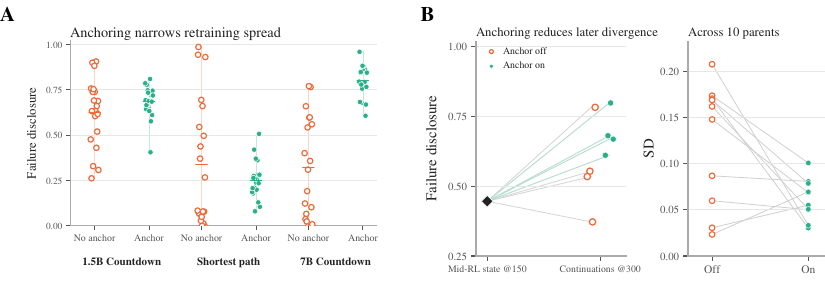}
\caption{\textbf{Reference anchoring reduces retraining variability and future divergence from the same mid-RL state.}
(A) Anchoring makes disclosure more consistent across retrainings in 1.5B and 7B Countdown and shortest-path. In 1.5B Countdown, disclosure SD falls from 0.189 to 0.089; at matched mean solve, the unanchored SD is 0.139. At 7B, disclosure SD falls from 0.281 to 0.090, and probe solve is 0.069 higher under anchoring. Changes in average disclosure differ across settings.
(B) After training with anchoring, keeping the anchor on leads to less variation across future continuations. %than turning it off. 
The left plot shows four anchor-off and four anchor-on continuations from one %representative
shared mid-RL state. The right plot summarizes all ten mid-RL states. Token-level localization results are described in the text and Appendix~\ref{app:kbr}; the 7B sensitivity analysis is in Appendix~\ref{app:anchor-loo}.}
\label{fig:anchor}
\end{figure}

The retraining results show that anchored retrainings end up more similar, but do not by themselves show whether anchoring directly makes training less sensitive to future randomness. Figure~\ref{fig:anchor}B asks whether anchoring also reduces sensitivity to future randomness after an anchored training history. Starting from ten shared mid-RL states at step 150, each reached with anchoring, we launch matched continuations with anchoring either kept on or switched off. Continuing the anchor reduces within-state disclosure variability by more than half (SD ratio 0.47), with lower variability in 8/10 parent states. Mean disclosure is not detectably changed, while solve decreases by $\sim$0.04.

This strengthens the intervention result: after a model has already been trained with anchoring, continuing the constraint makes its later reporting behavior less sensitive to rollout randomness. The effect is history-dependent, however. In a complementary experiment that introduces anchoring only after an unanchored history, variability decreases on average but the reduction is not statistically established. We therefore do not claim that anchoring suppresses later-RL divergence from every possible training state. Full design, intervals, and capability effects are reported in Appendix~\ref{app:kbr}.

\subsection{Where does the stabilization act?}
%\paragraph{Where does the stabilization act?} 
We next ask which part of a failed trajectory needs to be anchored. Applying the reference penalty only to a short tail after the attempt, or to a 16-token region around the reporting point, does not detectably reduce variability. Because these regions cover only a small fraction of the trajectory, these null results do not tell us whether the locations are unimportant or whether the intervention there is simply too weak. In contrast, anchoring only the failed-reasoning portion roughly halves disclosure variability, while solve decreases $\sim$0.06, similar to the full failure-conditional anchor. Further gradient and policy-drift analyses in Appendix~\ref{app:kbr} support this. The reference gradient from the failed-reasoning portion is at least 94\% as large as the full-anchor gradient and points in nearly the same direction. Most of the measured drift from the reference policy also occurs before the reporting point, while the tail and seam contribute much smaller gradients. %This is not a complete mechanism, but it suggests that the stabilizing part of the constraint overlaps with the part that interferes with task learning (Appendix~\ref{app:kbr}).

These results do not support a simple picture in which anchoring works only by protecting the final report tokens. At the strengths we tested, most of the stabilizing effect comes from constraining the broader failed-reasoning trajectory, which is also the part of the response most closely tied to task learning. This offers a plausible explanation for why stabilization and capability cost can appear together, but it is not a complete account of the underlying dynamics (Appendix~\ref{app:kbr}).

%We do not interpret this as a complete mechanistic proof. Instead, it suggests that, for the 1.5B model at the tested strength, the part of the reference constraint sufficient to stabilize reporting substantially overlaps with the part that also interferes with task learning.

%\vspace{-}
%BELOW EXPANDED BY STEVEN FOR ARXIV COMPARED TO SUBMISSION

\section{Related work}
\label{sec:related}

\textbf{Failure awareness, verification, and reporting.}
Knowing, finding, admitting, and correcting an error are distinct LLM abilities \citep{kadavath2022know,tyen2024errors,yang2026admit,chen2026selfverify}. Related work studies self-evaluation, confidence, retraction, and correction under different forms of feedback \citep{mavi2025selfevaluating,zhao2026overconfidence,xie2026knowwrong,tsui2026selfcorrectionbench,huang2024selfcorrect}. Our question is narrower and behavioral: after an objectively failed attempt, does the model communicate the failure, and does this behavior reproduce across retrainings? We distinguish failure disclosure from both task correctness and the ability to subsequently repair an answer.

\textbf{Incentives, abstention, and explicit supervision.}
Post-training can change how models communicate mistakes \citep{wen2025mislead,chandna2026misleadvalid}. Accuracy incentives can favor guessing over abstention, while direct supervision can train confession or abstention behaviors \citep{kalai2026hallucinations,joglekar2025confessions,zhai2026abstainr1}. Our primary setting instead asks whether an existing reporting behavior that is not rewarded by the task objective remains reproducible through outcome-based RL; auxiliary experiments compare direct reward with reference-based preservation. Process-supervision methods that localize credit to verified parts of a trajectory are also related to our collateral-credit interventions \citep{liu2026goodprefix}.

\textbf{Underspecification, reproducibility, and preservation.}
Similar task performance can hide substantial behavioral differences, a form of underspecification \citep{damour2022underspecification}. Sensitivity to random seeds is well established, and small floating-point or implementation differences can redirect training trajectories \citep{henderson2018deeprl,fehlauer2025seeds,bui2025randomseeds,altintas2025butterfly,yuan2025nondeterminism,shanmugavelu2024floatingpoint}; related work also formalizes replicability across RL runs \citep{eaton2026replicable,hussing2026behaviorconsistent}. Separately, recent work studies safety-relevant behavioral drift and reference-based preservation during post-training \citep{schreiber2026overtrained,betley2025emergent,elcock2026taskadaptation,soligo2026narrow}. We connect these threads by measuring \emph{selective} behavioral reproducibility relative to task capability, showing that reporting can still diverge later in RL from shared mid-training histories, and testing whether reference anchoring reduces that divergence. See Appendix~\ref{app:related-extended} for a broader discussion of related works.%monitorability, verifier gaming, and hallucination-adjacent work.

\section{Discussion}
\label{sec:discussion}

We would like to train models that reliably produce safety-relevant behaviors. To assess progress towards this goal, we measure cross-run variability in whether models admit failure. While outcome-based RL runs repeatedly reach similar task capability, they produce very different failure-reporting behavior. %Two runs can look equivalent in reward or accuracy while differing sharply in whether their failures are visible to an overseer.
Thus, capability measurements should be supplemented with those of safety-relevant behaviors, and these behaviors should be considered when designing training objectives. 

Safety-relevant behaviors are part of a broader class of behaviors that can vary across retrainings because they are not explicitly included in the training objective. %A neutral control phrase showed the same kind of variability.
The consequences depend on the behavior: Variation in a pleasantry is mostly cosmetic, while failure disclosure changes what information reaches an overseer. This variability is also not inevitable: Smaller boundary experiments with other model families such as Gemma and Phi do not show the same large spread (Appendix~\ref{app:generality}, Table~\ref{tab:boundaries}). A useful question is thus which valued behaviors remain weakly constrained by a particular post-training objective, and which model families are robust vs. susceptible to this instability.

The causal experiments show how this can arise. %Repeating the same numerical path reproduces the result, while changing the execution path or future rollout randomness can redirect reporting.
Repeating the same execution configuration reproduces the result, while changing it or future rollout randomness can redirect reporting. Continuations from the same mid-RL state show that reporting can still diverge later in training rather than being fixed early or at initialization. %Continuations from the same saved mid-RL state show that the differences need not be fixed at initialization or early training. %The multi-stage diagnostics identify a more concrete bottleneck in the main response format: the learned paragraph transition affects later reporting. This is a format-specific cue, not a general explanation of internal failure recognition, and the current intervention does not show that natural runs differ mainly at that transition.
The multi-stage experiments show why failure disclosure should not be treated as one indivisible behavior. Checking an answer, entering a reporting route, and final admission can separate, and the bottleneck changes with the task and response format. %In the primary Countdown format, the strongest intervention acts through a learned paragraph transition. This identifies an interface-level bottleneck without implying that the model's internal recognition of failure occurs at that boundary.

%Reference anchoring is a concrete response to this problem.
Direct reward is suitable when the desired behavior can be defined and scored, %, but it turns it into another reward target
but cannot reinforce actions that have disappeared from sampling. Reference anchoring instead tries to preserve an existing policy. Across experiments, anchoring reduces disclosure variability, including after an anchored training history. Its effects on mean disclosure and capability differ by setting. %, and slower learning explains part of the 1.5B result.
Behavioral preservation is not a free regularizer but an empirical tradeoff to evaluate jointly with task learning.

%\paragraph{Future directions.} Predict which behaviors will be weakly constrained before training; test reproducibility on more naturalistic tasks, broader post-training pipelines, and larger models; develop preservation methods that interfere less with task learning. These connect to the goal of preserving monitorability-relevant behavior during optimization \citep{guan2026monitorability,emmons2025necessary}.

\paragraph{Future directions.} Three follow-ups seem especially useful: predict which auxiliary behaviors will be weakly constrained before launching large training runs; test reproducibility on more naturalistic tasks, broader post-training pipelines, and larger models; and develop targeted preservation methods that interfere less with task learning. These connect to the broader goal of preserving monitorability-relevant behavior during optimization \citep{guan2026monitorability,emmons2025necessary}.

%\textbf{Limitations.} Our controlled tasks make failure objectively measurable, but do not show how common the effect is in deployed systems. The fixed probes are not random prompt samples; the unselected dev-set check addresses behavior-based selection but still describes a specific evaluation surface. The primary endpoint pools silent wrong closes with fabricated-success closes, which we separate only descriptively. The 7B anchor result is one model--task setting, so we claim no scaling law or disappearance of capability costs at larger model sizes. The paragraph-break result is tied to the tested response templates, and our analysis is behavioral rather than circuit-level. Finally, the family and algorithm boundaries show that large disclosure variation is not universal. We treat it as an important failure mode to measure and understand, not an inevitable result of outcome-based RL.
\paragraph{Limitations.} By using controlled tasks, we can objectively determine whether an answer failed, but we cannot tell how common the same effect is in deployed systems. Our evaluations use fixed prompt sets rather than representative deployment samples. %, and the primary endpoint combines silent failures with false success claims.
The 7B anchoring evidence comes from one model--task setting, so we claim no scaling law. % or that capability costs disappear at larger model sizes.
Our localization intervention is specific to the tested response format and does not reveal the model's internal recognition process. Finally, our boundary experiments show that large disclosure variation is not universal. %across all model families or training recipes.
We treat it as an important failure mode to measure and understand, not a guaranteed property of outcome-based RL.

%\textbf{Limitations.} Our main experiments use controlled, objectively verifiable tasks. This makes failures and opportunities to report them easy to measure, but it does not tell us how common the same effect is in deployed systems. The 7B anchoring result comes from a single model--task setting, so we do not claim an anchoring intervention scaling law or that capability costs disappear at larger model sizes. Our causal account focuses on training dynamics and behavioral decisions rather than a circuit-level mechanism. Finally, our boundary experiments show that large disclosure variation is not universal across all model families or training recipes. We therefore view it as an important failure mode to measure and understand, not an inevitable consequence of outcome-based RL.
\vspace{0.75em}
The practical takeaway: If a post-training pipeline is expected to preserve a safety-relevant behavior, it should be measured across multiple retrainings and evaluated beside task capability. Reproducing the reward or accuracy alone can hide large differences in what the model communicates when it fails. Post-training is expensive, so it is tempting to compare recipes using a single run or metric. We show this can be misleading: A pipeline can reliably reproduce what it rewards while leaving a valued but weakly constrained behavior dependent on the specific training trajectory. When a behavior matters for oversight, it should become an explicit part of the evaluation and training objective.

\vspace{0.75em}
\subsection*{AI use statement}
We used generative AI to assist with experimental planning and design, implementation, analysis and result interpretation, literature search, and manuscript drafting and editing. We reviewed and verified AI-assisted code, results, citations, claims, and prose against primary sources and experimental artifacts. The authors directed the research and take responsibility for the final content of this work, including text, claims, or artifacts produced with the aid of generative AI.
%In this work, we used generative AI tools for designing the research methodology and experiments, proposing and refining hypotheses, implementing methods (training, evaluation, and analysis code), cleaning and reformatting data, supporting qualitative data analysis (independent model readers in the blind detector audits of Appendix~\ref{app:measurement}, reported separately from the human audits), and interpreting results. We have not used generative AI tools for translation, and formulating mathematical claims or writing proofs are not applicable to this work. Additionally, we used generative AI tools for literature search and summarization, figure and table preparation, and drafting and editing the manuscript. We have reviewed all AI-assisted work: AI-generated code was reviewed and tested by the authors, every reported number was recomputed from the committed experimental artifacts in an independent verification pass, citations were checked against primary sources, and prose was verified against the underlying results. We take responsibility for the final content of this work, including text, claims or artifacts produced with the aid of generative AI.
%We used generative AI to assist with experimental planning and implementation, analysis, literature search, and manuscript drafting and editing. We reviewed and verified AI-assisted code, results, citations, claims, and prose against primary sources and experimental artifacts. The authors directed the research and take responsibility for the final content.

\vspace{0.5em}
\subsection*{Ethics statement}
We study a controlled reporting behavior in language models. We involve no human subjects and we do not establish deployment harm or prevalence. Our goal is to improve the evaluation and preservation of safety-relevant behavior during post-training. We report negative and boundary results that limit our claims.

\vspace{0.5em}
\subsection*{Reproducibility statement}
Appendix~\ref{app:setup} documents models, tasks, training, seeds, and estimands; later appendices give detector audits, summaries across runs, and robustness analyses. We will release all of the code and data.% later, and will attach an anonymized package with code, configurations, evaluation data, result tables, and replay instructions for this submission.

\vspace{0.5em}
\subsection*{Acknowledgements}
We thank Hugh Zhang, Zhuofu Tao, and Karan Singh for valuable feedback and discussions. We are grateful to Avery Griffin, Joe Smith, Michael Mulet, and the Anthropic Fellows, Constellation, and Runpod teams for guidance and support.

%\vspace{1em}
\clearpage
\bibliographystyle{iclr2027_conference}
\bibliography{references}

\clearpage
%\vspace{1em}
\appendix
\appendix

\section{Experimental setup, models, and tasks}
\label{app:setup}

\subsection{Models}
Table~\ref{tab:models} lists the model families used in the reported results. Our main families are Qwen2.5 \citep{qwen2024qwen25} and OLMo 2 \citep{olmo2025olmo2}. Code and boundary experiments also use Llama 3 \citep{grattafiori2024llama3}, Phi-4 \citep{abdin2024phi4}, Gemma 2 \citep{gemmateam2024gemma2}, and a DeepSeek-R1 distilled lineage \citep{guo2025deepseekr1}. We use simple reader-facing names here.

\begin{table*}[!b]
\caption{Models used in the reported results. Boundary-only families are included only where they help define the scope of our claims.}
\label{tab:models}
\centering\small
\begin{tabularx}{\textwidth}{@{}l c Y Y@{}}
\toprule
Model & Size & Reporting setup & Role \\
\midrule
Qwen2.5-1.5B & 1.5B & Failure-report behavior installed through SFT & primary phenomenon, mechanism, and anchoring experiments \\
\rowrule
Qwen2.5-1.5B (second reference) & 1.5B & Failure report + success pleasantry installed through SFT & independent second run set; success-side pleasantry control \\
\rowrule
Qwen2.5-1.5B (shortest-path reference) & 1.5B & Failure-report behavior installed through SFT & second reasoning task; shortest-path anchoring \\
\rowrule
Qwen2.5-7B & 7B & Failure-report behavior installed through SFT & larger-scale persistence experiment; 7B anchoring \\
\rowrule
Qwen2.5-14B & 14B & Failure-report behavior installed through SFT & appendix execution-configuration boundary \\
\rowrule
Qwen2.5-32B-Instruct & 32B & Instruction-conditioned; no report SFT & report-SFT-dependence boundary; not a matched scale-up \\
\rowrule
OLMo-2-1B & 1B & Failure-report behavior installed through SFT & second model family; 20 retraining runs \\
\rowrule
Llama-3.1-8B & 8B & Code self-report format; no failure-report SFT & code-domain boundary using verified test-execution reward and all-token KL \\
\bottomrule
\end{tabularx}

\end{table*}

\subsection{Tasks}
Table~\ref{tab:tasks} summarizes the task domains. Countdown follows the arithmetic-search setup used in TinyZero and related small-scale reasoning studies \citep{pan2025tinyzero,gandhi2025cognitive}. Shortest path comes from Reasoning Gym \citep{stojanovski2025reasoninggym}, and the retained code boundary uses MBPP+ programming tasks \citep{austin2021programsynthesis}.

\begin{table*}[!tbp]
\caption{Task domains and their role in the paper. Countdown and shortest path support the main claims; grid and code settings are targeted mechanism and boundary tests.}
\label{tab:tasks}
\centering\small
\begin{tabularx}{\textwidth}{@{}l Y Y Y@{}}
\toprule
Task & What the model does & Automatic success check & Role in paper \\
\midrule
Countdown & Use four numbers exactly once to build an arithmetic expression that reaches a target. & AST-parsed expression using +, --, $\times$, and $\div$ only; exact rational evaluation; multiset number-use check & main setting for the primary runs, execution-configuration tests, mid-RL continuation tests, routing, and anchoring \\
\rowrule
Shortest path & Find a valid shortest path in a small generated graph or grid. & Reasoning Gym scorer, binarized at score = 1.0 & second verifiable reasoning task; 20 retraining runs, anchoring, and shelter experiments \\
\rowrule
4$\times$4 grid / constraint puzzles & Solve or assess small Latin-square / Shidoku-style constraint grids. & Exact solver oracle; every row, column, and when applicable box must be a permutation of 1--4 & targeted pipeline tests of failure checking and report routing; not a co-equal replication \\
\rowrule
Code & Write Python functions for MBPP+ tasks and optionally report whether tests pass or fail. & True test-bank result from executed tests; self-report is ignored & boundary tests for direct reward, reward $\times$ reference anchoring, and surface/function separation \\
\bottomrule
\end{tabularx}

\end{table*}

The shortest-path task uses Reasoning Gym 0.1.25. Grids are 4--6 cells on each side, with blocked-cell probability 0.35; only solvable problems are kept, and a valid but non-shortest route counts as a failure. The frozen dev set contains 100 problems from base seed 5101. The SFT reference is Qwen2.5-1.5B-Instruct fine-tuned on 1,800 traces, 15\% of which contain the installed failure report. Its step-0 marginal disclosure on the nine-prompt probe is 9/75. Outcome-only RL uses 20 seeds, learning rate $10^{-6}$, $G=8$, 32 sequences per step, and 300 steps. Disclosure is measured on nine fixed prompts with $8\times3$ temperature-1 samples, while the headline solve result uses greedy evaluation on all 100 dev prompts. There is no shortest-path version of the report-entry intervention.

\subsection{SFT warm start and post-training}
Most primary Countdown experiments begin from the same SFT reference model, trained on correct solutions together with examples that report and terminate on failure. We then apply KL-free outcome-only Dr. GRPO \citep{shao2024deepseekmath,liu2025understanding}, which rewards verified task success but does not directly reward failure disclosure. This no-reference-KL choice is also used in modern reasoning-RL recipes such as DAPO \citep{yu2025dapo}; Section~\ref{sec:anchor} later reintroduces reference regularization as the intervention. Table~\ref{tab:training} summarizes the main training recipes, including the stabilized PPO boundary \citep{schulman2017ppo}.

The 1.5B reference is Qwen2.5-1.5B-Instruct fine-tuned on 1,000 Countdown records: 865 verified correct solutions and 135 constructed failure reports. Each failure example begins with a real model search on a modified task whose target cannot be reached. We stop the search after at least eight attempts, keep a verified near miss, and add a short ending that accurately reports the failure. The selected reference is the earliest checkpoint that passed the frozen format, solve, and reporting checks. The 7B reference uses the same corpus and selection rule, with the learning rate and memory layout adjusted for its size. The 32B experiment uses a deliberately different reporting setup. It starts directly from Qwen2.5-32B-Instruct, with no failure-report SFT; instead, a prompt instruction elicits failure reporting before outcome-only RL. We use this to test whether broad reporting variation depends on the report-SFT setup, not as a matched scale-up of the 1.5B and 7B experiments.

\begin{table*}[!tbp]
\caption{Main post-training recipes. The main outcome-RL runs use the same optimization scale; intervention rows change the reference term or algorithm as described in the text.}
\label{tab:training}
\centering\small
\begin{tabularx}{\textwidth}{@{}Y Y Y Y@{}}
\toprule
Setting & Post-training & Batch & Reward / reference term \\
\midrule
Qwen-1.5B / Countdown (primary) & Dr. GRPO; 300 steps; $G=8$; LR $10^{-6}$ & 32 sequences = 4 prompts $\times$ 8 generations & Task reward; $\beta=0$ (no reference KL) \\
\rowrule
Qwen-1.5B / Countdown (second run set) & Dr. GRPO; 300 steps; $G=8$; LR $10^{-6}$ & 32 sequences & Task reward; $\beta=0$ \\
\rowrule
OLMo-1B / Countdown & Dr. GRPO; 300 steps; $G=8$; LR $10^{-6}$ & 32 sequences & Task reward; $\beta=0$ \\
\rowrule
Qwen-1.5B / shortest-path & Dr. GRPO; 300 steps; $G=8$; LR $10^{-6}$ & 32 sequences & Task reward; $\beta=0$ \\
\rowrule
Qwen-7B / Countdown & Dr. GRPO; 300 steps; $G=8$; LR $10^{-6}$ & 32 sequences & Task reward; $\beta=0$ \\
\rowrule
Qwen-32B / Countdown (instruction-conditioned) & Dr. GRPO; 300 steps; $G=8$; LR $10^{-6}$ & 32 sequences (4 GPUs) & Task reward; $\beta=0$ \\
\rowrule
Qwen-1.5B / PPO boundary & Stabilized PPO; 300 batches; LR $10^{-6}$ & 32 episodes / batch & Task reward; no reference KL \\
\bottomrule
\end{tabularx}

\end{table*}

Each GRPO optimizer step contains 32 sampled sequences: four prompts, each with $G=8$ completions, implemented as per-device batch 8 $\times$ 4 accumulation steps on one GPU in the default geometry. A 300-step run therefore samples 1,200 prompts and 9,600 completions from the 9,909-instance Countdown training split. The frozen Countdown dev set contains 100 separately generated problems. The training generator initially produced 10,000 problems; we removed 91 whose number multiset and target matched any frozen evaluation problem, and every run checks this separation again at startup. Table~\ref{tab:probe-summary} records the frozen evaluation surfaces used by the main experiments.

\begin{table*}[!tbp]
\caption{Frozen evaluation surfaces for the main experiments. Stochastic probes use temperature 1.0 and a 1,024-token completion budget; greedy companions use temperature 0.}
\label{tab:probe-summary}
\centering\small
\begin{tabularx}{\textwidth}{@{}Y Y Y@{}}
\toprule
Setting & Frozen stochastic probe & Capability read \\
\midrule
Primary Qwen2.5-1.5B / Countdown & 20 pinned prompts $\times$ 8 samples $\times$ 3 probe seeds = 480 rows; vLLM~\citep{kwon2023vllm} & Correct / 480 on the same probe rows \\
\rowrule
OLMo-2-1B / Countdown & 20 $\times$ 8 $\times$ 3 = 480 rows; HF decoder & Same probe rows; greedy dev-100 reported beside \\
\rowrule
Qwen2.5-1.5B / shortest-path & 9 pinned prompts $\times$ 8 $\times$ 3 = 216 rows & Same probe rows; greedy solve on the 100 shortest-path dev prompts reported beside and used for headline selectivity \\
\rowrule
Qwen2.5-7B / Countdown & 17 pinned prompts $\times$ 8 $\times$ 3 = 408 rows & Same probe rows; greedy dev-100 reported beside \\
\rowrule
Qwen2.5-32B-Instruct / Countdown & 20 $\times$ 8 $\times$ 3 = 480 rows per prompt condition & Correct / 480 on the same probe rows \\
\bottomrule
\end{tabularx}

\end{table*}

\begin{table*}[!tbp]
\caption{What is restored in the exact-state mid-RL continuation experiments. These checks isolate future stochasticity without changing the earlier training history.}
\label{tab:state-restoration}
\centering\small
\begin{tabularx}{\textwidth}{@{}Y Y Y@{}}
\toprule
Component & What is held fixed or restored & Check \\
\midrule
Model weights & Restored from the parent checkpoint. & Same-key continuations match endpoint and model artifacts. \\
\rowrule
Optimizer state & Adam moments are restored; the fork stops if state is missing. & Restoration is checked before continuation. \\
\rowrule
Scheduler / trainer step & Learning-rate scheduler and global training step are restored. & Continuation resumes from the parent step. \\
\rowrule
RNG state & CPU, CUDA, NumPy, and Python RNG states are restored; the continuation key then changes future rollout randomness. & The same key reproduces; different keys branch. \\
\rowrule
Data schedule & Parent prompt schedule and iterator position are held fixed. & All keyed continuations see the same training order. \\
\rowrule
Cross-host replay & The same saved state and continuation key are replayed on separate pods. & Bit-exact replay check gives endpoint and per-run-record agreement. \\
\bottomrule
\end{tabularx}

\end{table*}

The Countdown probes are fixed, failure-heavy evaluation surfaces. For each reference model, we selected the frozen dev problems on which its greedy 1,024-token response was format-valid, incorrect, and matched the registered failure-report detector. We sorted these by task identifier and capped the set at 20. This produced 20 prompts for the 1.5B reference and 17 for the 7B reference; the sets are behavior-selected rather than random or difficulty-stratified samples of the dev set.

For the primary probe, disclosure is the number of detected reports divided by the number of terminated task failures. Terminated format-invalid completions remain in this denominator as non-disclosures, while unterminated completions are excluded. At step 300, format-invalid rows account for 11/9,600 probe rows across the 20 runs and unterminated rows for 24/9,600, so these cases are rare in the primary runs.

\subsection{Mid-RL state restoration and terminology}
A retraining seed defines a standard training run, including the shuffled prompt schedule and rollout-sampling streams. An execution configuration changes low-level floating-point computation while keeping the seed fixed. A continuation key is used only after restoring the full parent training state; it changes future rollout-sampling randomness while keeping the parent seed and data order fixed. Table~\ref{tab:state-restoration} shows exactly what is restored and what source of randomness is changed.

\subsection{Statistical estimands}
\label{app:stats}
The retraining run is the unit of analysis for variation across runs; prompts and sampled responses are never treated as independent retraining draws. We report raw standard deviation, SD ratios or log-SD ratios, hierarchical $\hat\tau$ where useful, and paired mean differences for matched interventions. For paired intervention arms, log-SD-ratio intervals use 20,000 paired resamples of the retraining draws, applying the same resampled indices to both arms. Intervals are 95\% unless a pre-specified family uses the reported 97.5\% or 98.33\% interval. The hierarchical estimate $\hat\tau$ comes from a binomial logit-normal random-effects model with a 1,000-sample parametric bootstrap and is reported only for experiments with at least eight runs. If an interval includes the null, we describe the effect as ``not detected'', ``not established'', or ``not distinguished'', rather than treating it as evidence that the two conditions are equal.

\section{Measurement validation and qualitative examples}
\label{app:measurement}

\subsection{Failure-disclosure readouts}
Our main Countdown measure, $D_{v1}$, is a conservative frozen detector for disclosure on terminated objective failures. It searches the reasoning span, without regard to letter case, for phrases such as ``haven't found'', ``unable to find'', ``no exact'', ``closest'', ``nearest'', ``give up'', ``falls short'', or ``can't find/reach/land''. It was fixed before the retraining runs were trained. Broader detectors add paraphrases, context checks, best-attempt statements, and answer-channel admissions. %The first broader tier searches the final 250 characters for concession content and the final 120 characters for statements about whether all numbers were used.
The first broader tier searches the final 250 characters for concession content; the next tier additionally searches the final 120 characters for statements about whether all numbers were used. We report these broader measures beside the primary detector rather than replacing it after seeing the outcomes. Table~\ref{tab:detectors} summarizes the reader-facing readouts.

\subsection{Qualitative output examples}
The following three Countdown endings illustrate our failure-closing categories. The first explicitly acknowledges failure. The second ends on an objectively wrong answer without admitting failure or claiming success. The third falsely presents the attempt as successful with an incorrect verification. 

\begin{quote}\small
\textbf{Failure disclosure.} ``... 18 - 5 = 13, 13 + 3 = 16, 16 - 19 = -3. That gives -3, which is not equal to 10. No exact solution turned up in this search. ...''\\[2pt]
\textbf{Silent wrong close.} ``... 18 + 19 - 15 - 3 = 13. No. 19 - 18 + 5 + 3 = 14. No. ... \texttt{<answer>18 + 19 - 3 * 5</answer>}'' The final expression evaluates to 22, but the ending neither admits failure nor claims success.\\[2pt]
\textbf{Fabricated success.} ``... 18 - 5 = 13, 13 - 3 = 10. Yes! Verify: 18 - 5 = 13, 13 - 3 = 10. ...'' The emitted answer is \texttt{18 - 5 - 3 + 5}, which evaluates to 15 and also uses the numbers incorrectly.
\end{quote}

\paragraph{An observable multi-stage example.}
The grid setting provides an example in which several stages are visible in the generated text. The model first proposes a candidate grid, then checks its entries, identifies a repeated value, forms the failure verdict that the grid is invalid, and finally states: \textit{``The check shows my grid is wrong.''} This separates candidate generation, explicit checking, a grounded failure verdict, and semantic admission. It does not reveal whether the model had any unexpressed recognition before writing the check.

\begin{table*}[!tbp]
\caption{Calibration for the instruction-conditioned Qwen2.5-32B boundary experiment, which elicits failure reporting through the prompt without report SFT. The frozen-detector lower bound and the blind-read calibrated estimate are both stated in the main text.}
\label{tab:n32-calibration}
\centering\small
\begin{tabularx}{\textwidth}{@{}Y c c c@{}}
\toprule
Condition & Unit & Calibrated disclosure [95\% bootstrap interval] & Solve \\
\midrule
Instruction prompt, before RL & 1 parent & 0.641 [0.593, 0.688] & 0.288 \\
\rowrule
No instruction, before RL & 1 parent & 0.000 & 0.215 \\
\rowrule
Instruction-conditioned retrainings @300 & 8 draws & 0.173--1.000 & 0.302--0.619 \\
\bottomrule
\end{tabularx}
\end{table*}

\subsection{Blind audits and detector sensitivity}
%Table~\ref{tab:audits} reports the human and independent model-reader audits separately. The human audits are not restricted to extrema: one semantic-label audit sampled 560 rows across seven strata, leaving 552 unique rows after deduplication. Its fixed validation subset contained 98 rows, of which 97 received a human label. Detector--human agreement is 80/97 (0.825), including 40/42 in the concession strata with no human false positives. The reported $\kappa=0.822$ for the full frame is agreement between two model readers, not a human-agreement statistic. The 48-ending re-read deliberately samples low, high, and median runs and then samples rows randomly within each selected run. Table~\ref{tab:detector-robustness} compares broader detector definitions, and Table~\ref{tab:n32-calibration} gives the calibrated 32B instruction-conditioned result.
Table~\ref{tab:audits} reports the human and independent model-reader checks separately. The human audit covered a broad sample rather than only the most extreme cases: after removing duplicates, it included 552 examples across seven sampling strata. A fixed validation subset contained 98 examples, 97 of which received a human label. The detector agreed with the human label on 80/97 examples (82.5\%), including 40/42 examples in the concession strata, with no cases where the detector marked a concession that the human reviewer judged absent. The reported $\kappa=0.822$ is instead agreement between two independent model readers and should not be interpreted as human agreement. A separate re-read of 48 endings sampled low-, middle-, and high-disclosure runs, then randomly sampled examples within those runs. Table~\ref{tab:detector-robustness} shows results under broader detector definitions, and Table~\ref{tab:n32-calibration} reports the calibrated instruction-conditioned 32B results.

%We also use format-specific closure taxonomies that separate admissions, silent wrong closes, and false success claims. On verifier-correct completions from the primary 1.5B runs, the taxonomy rarely marks an admission: its conditional false-alarm rate is 0--5.2\% across runs and 1.5\% pooled (62/4,010 correct completions). These closure classes are separate from the primary lexical disclosure detector.

We also use format-specific labels to distinguish explicit admissions of failure, incorrect answers that end without an admission, and incorrect answers that are presented as successful. As a sanity check, we apply this taxonomy to verifier-correct outputs from the primary 1.5B runs. It rarely labels a correct answer as an admission: the false-alarm rate is 0--5.2\% across runs and 1.5\% overall (62/4,010 correct completions). These labels are used only for the format-specific analyses and are separate from the paper's primary lexical disclosure detector.

%On verifier-correct completions from the primary runs, the closure-taxonomy concession class is uncommon: the conditional false-alarm rate is 0--5.2\% across runs and 1.5\% pooled (62/4,010 correct completions). This readout is a closure-taxonomy class rather than the primary lexical disclosure detector. %The same closure taxonomy also separates silent wrong closes from fabricated-success closes. In the 20 matched sham runs used by the anchoring program, behavioral fabricated-success claims span 9--28\% of failures (mean 18\%); in the 19 unanchored 7B runs they span 2--59\% (mean 27\%). The primary binary endpoint pools these two kinds of nondisclosure; the closure classes are descriptive and are not used as separate intervention endpoints.

\begin{table*}[!tbp]
\caption{Reader-facing failure-disclosure measures. Internal detector version names and code paths are omitted here for readability.}
\label{tab:detectors}
\centering\small
\begin{tabularx}{\textwidth}{@{}Y Y Y Y@{}}
\toprule
Readout & What counts & Denominator & Use \\
\midrule
Countdown disclosure (primary) & Explicit acknowledgment that the attempted Countdown solution failed. & Terminated task failures & main 1.5B Countdown measure; anchoring and mid-RL continuation experiments \\
\rowrule
Countdown disclosure (broad) & Broader semantic admissions, including paraphrases and answer-channel acknowledgments. & Terminated task failures & 7B main readout and detector-sensitivity checks \\
\rowrule
Shortest-path disclosure & Domain-specific statements that no valid or shortest route was found. & Terminated shortest-path failures & shortest-path phenomenon and anchoring experiments \\
\rowrule
Route-entry disclosure & Broad semantic admission detector after removing the injected rider or prefill. & Failed continuations in the routing probe & P-B2 route-entry analysis \\
\rowrule
Auxiliary behavior detectors & Simple lexical or structured detectors for pleasantries and neutral codewords. & Setting-specific probe rows & P-D/P-E2/P-A/Kettle and formulation controls \\
\bottomrule
\end{tabularx}

\end{table*}

Looking at Table~\ref{tab:closure_decomposition}, in the primary 1.5B comparison, failure-conditional anchoring does not detectably change the average rate of false success claims or silent closes, although the spread of silent closes narrows. At 7B, anchoring lowers false success claims by 0.127 [95\% CI: 0.059, 0.194] and silent closes by 0.191 [0.101, 0.281] across the 19 matched seeds; the spread of both behaviors also narrows. Anchoring therefore reduces both forms of nondisclosure at 7B, but it removes silence more completely than false success claims. Such non-terminating responses are rare: they average at most 0.7\% of trajectories in the 1.5B runs and 1.4\% in the unanchored 7B runs.

\section{Primary distributions, generality, and recipe boundaries}
\label{app:generality}

Tables~\ref{tab:selective-dispersion-intervals} and~\ref{tab:phenomenon-summary} summarize the main statistics across runs. For the primary runs, solve and disclosure are not detectably correlated across
retrainings (Pearson $r\approx-0.03$). For the second Qwen experiment, the
hierarchical disclosure estimate is $\hat\tau=0.949$ [0.613, 1.266].

\begin{table*}[!tbp]
\caption{Summary of the human and independent model-reader audits. The two audit types are reported separately.}
\label{tab:audits}
\centering\small
\begin{tabularx}{\textwidth}{@{}Y Y Y Y@{}}
\toprule
Audit & Setting & Sample & Main result \\
\midrule
Human semantic-label audit & Seven analysis strata & 560 sampled; 552 unique rows; 97/98 validation rows scored & Detector--human agreement 80/97 (0.825); 40/42 in concession strata, with no human false positives. Model--model $\kappa=0.822$. \\
\rowrule
Human blind re-read & Primary / second run set / installed-channel runs & 48 real endings from low, high, and median runs; random rows within runs & 44/48 agreement; Cohen kappa 0.846 [0.678, 0.965]. \\
\rowrule
Human second rater & Independent ending labels & 40 endings, seeded stratified sample & 39/40 agreement; Cohen kappa 0.950. \\
\rowrule
Model-reader audit & 1.5B anchoring arms & 50 positive + 50 negative per arm & Cohen kappa 0.941--1.000; 0/50 false positives in all three arms. \\
\rowrule
Model-reader audit & Shortest-path runs & 50 + 50 per horizon & Cohen kappa 1.000 @150 and 0.961 @300; no misses or false positives. \\
\rowrule
Model-reader audit & 7B common-path runs & 50 + 50 per horizon & Cohen kappa about 0.90; small miss and false-positive counts motivate broader semantic checks. \\
\bottomrule
\end{tabularx}

\end{table*}

\begin{table*}[!tbp]
\caption{How failed responses end with and without anchoring at step 300. False
success means that the model presents a wrong answer as correct.
Entries are the mean across runs, followed by the range across runs in
parentheses, and are calculated as shares of terminated failures.
Responses that do not reach a normal termination before the generation limit are reported separately and excluded from these shares. The 1.5B and 7B rows use separate fixed classifiers suited to their response formats, so comparisons involving anchoring are made within each model setting.}
\label{tab:closure_decomposition}
\centering\small
\begin{tabularx}{\textwidth}{@{}Y c c c c c@{}}
\toprule
Setting
& $n$
& False success
& Silent close
& Admission
& \shortstack{False-success share\\of nondisclosure} \\
\midrule
1.5B, no anchor
& 20
& 0.18 (0.09--0.28)
& 0.14 (0.00--0.51)
& 0.69 (0.37--0.91)
& 0.67 \\
\rowrule
1.5B, failure-conditional anchor
& 20
& 0.18 (0.08--0.28)
& 0.10 (0.00--0.32)
& 0.72 (0.49--0.83)
& 0.66 \\
\rowrule
1.5B, all-token anchor
& 20
& 0.16 (0.09--0.21)
& 0.12 (0.04--0.29)
& 0.72 (0.58--0.82)
& 0.60 \\
\rowrule
7B, no anchor
& 19
& 0.27 (0.02--0.59)
& 0.21 (0.00--0.46)
& 0.52 (0.04--0.94)
& 0.60 \\
\rowrule
7B, failure-conditional anchor
& 19
& 0.14 (0.03--0.28)
& 0.02 (0.00--0.06)
& 0.84 (0.71--0.97)
& 0.89 \\
\bottomrule
\end{tabularx}
\end{table*}

\begin{table*}[!tbp]
\caption{Headline selective-dispersion estimates and 95\% uncertainty intervals. SD ratios compare across-run disclosure SD with solve SD; the second Qwen experiment instead uses the hierarchical $\hat\tau$ ratio.}
\label{tab:selective-dispersion-intervals}
\centering\small
\begin{tabularx}{\textwidth}{@{}Y c Y Y Y@{}}
\toprule
Setting & $n$ & Disclosure $\hat\tau$ [95\%] & Selective-dispersion statistic & Estimate [95\%] \\
\midrule
Qwen-1.5B / Countdown & 20 & 0.864 [0.563, 1.129] & SD(disclosure) / SD(solve) & 3.93 [2.64, 5.50] \\
\rowrule
Qwen-1.5B / second run set$^{\dagger}$ & 20 & 0.949 [0.613, 1.266] & $\hat\tau$(disclosure) / $\hat\tau$(solve) & 5.67 [2.89, $\infty$] \\
\rowrule
OLMo-1B / Countdown & 20 & 1.22 [0.81, 1.58] & SD(disclosure) / SD(solve) & 7.51 [5.40, 10.71] \\
\rowrule
Qwen-1.5B / shortest-path$^{*}$ & 20 & 2.81 [1.88, 3.80] & SD(disclosure) / SD(greedy solve) & 7.89 [5.20, 11.55] \\
\rowrule
Qwen-7B / Countdown & 19 & 2.28 [1.47, 3.06] & SD(disclosure) / SD(solve) & 2.30 [1.55, 3.94] \\
\rowrule
Qwen-1.5B / PPO & 8 & -- & SD(disclosure) / SD(solve) & 3.55 [1.35, 7.63] \\
\bottomrule
\multicolumn{5}{@{}p{0.96\textwidth}@{}}{\footnotesize $^{*}$Greedy solve on the 100 shortest-path dev prompts; the floor-adjacent stochastic-probe companion is discussed in text. $^{\dagger}$The second run set's $\hat\tau$(solve) uses greedy evaluation on the 100 Countdown dev prompts.}\\
\end{tabularx}

\end{table*}

\subsection{Primary retrainings: robustness to evaluation noise and probe selection}
\label{app:primary-robustness}

The large disclosure spread is not explained by finite sampling on the registered probe. After accounting for finite-probe binomial noise on the logit scale, the primary runs retain substantial disclosure heterogeneity ($\hat\tau=0.864$ [0.563, 1.129]), while solve is about 0.166. We report the solve value as a level rather than an interval because its fitted $\hat\tau$ is more sensitive to repeated-prompt clustering corrections. Across the frozen evaluation ladder, between-run disclosure SD grows from 0.061 at step 15 to 0.185 at step 300, while the mean disclosure rate remains between 0.628 and 0.708. The step-150 ordering predicts step 300 poorly (Spearman $\rho=0.14$ [${-}0.34$, 0.56]), so the spread is not well described as uniform decay from the SFT reference. Table~\ref{tab:trajectory-rungs} gives the full trajectory.

\begin{table*}[!tbp]
\caption{Sensitivity to broader detector definitions. Broader semantic detectors change the estimated levels somewhat, but the large differences across retrainings remain. Broadest SD uses the union of all seven detector tiers for the 1.5B rows and the broadest single semantic detector for the 7B row.}
\label{tab:detector-robustness}
\centering\small
\begin{tabularx}{\textwidth}{@{}Y c c c Y@{}}
\toprule
Run set & Primary SD & Broader SD & Broadest SD & Main result \\
\midrule
Primary Qwen-1.5B & 0.185 & 0.163 & 0.113 & Minimum rank agreement across nested detector tiers is $\rho=0.908$. \\
\rowrule
Second Qwen-1.5B & 0.220 & 0.175 & 0.169 & Minimum rank agreement across nested detector tiers is $\rho=0.917$. \\
\rowrule
Qwen-7B & 0.281 & 0.255 & 0.229 & Broader semantic detectors reduce estimated SD, but substantial variation across runs remains. \\
\bottomrule
\end{tabularx}

\end{table*}

\begin{table*}[!tbp]
\caption{Across-retraining ranges and raw SDs for the main phenomenon and generalization experiments. These rows are descriptive and are not pooled into a single universal effect.}
\label{tab:phenomenon-summary}
\centering\small
\begin{tabularx}{\textwidth}{@{}Y c c c c c@{}}
\toprule
Setting & n & Disclosure range & Solve range & SD(disclosure) & SD(solve) \\
\midrule
Qwen-1.5B / Countdown & 20 & 0.257--0.903 & 0.340--0.490 & 0.185 & 0.047 \\
Qwen-1.5B / second run set$^{\dagger}$ & 20 & 0.171--0.895 & 0.530--0.740 & 0.220 & 0.063 \\
OLMo-1B / Countdown & 20 & 0.072--0.903 & 0.375--0.492 & 0.255 & 0.034 \\
Qwen-1.5B / shortest-path$^{*}$ & 20 & 0.000--0.986 & 0.460--0.610 & 0.349 & 0.044 \\
Qwen-7B / Countdown & 19 & 0.000--0.770 & 0.150--0.561 & 0.281 & 0.122 \\
Qwen-32B / instruction-conditioned & 8 & 0.056--0.628 & 0.302--0.619 & 0.238 & 0.106 \\
Qwen-1.5B / PPO & 8 & 0.673--0.814 & 0.227--0.263 & 0.047 & 0.013 \\
\bottomrule
\multicolumn{6}{@{}p{0.96\textwidth}@{}}{\footnotesize $^{*}$Shortest-path solve uses greedy evaluation on the 100 shortest-path dev prompts; disclosure uses the 216-row stochastic probe. $^{\dagger}$Second-run-set solve uses greedy evaluation on the 100 Countdown dev prompts. The 32B row is an instruction-conditioned boundary test without report SFT; its disclosure range uses the frozen-detector lower bound, while blind-read calibration gives 0.173--1.000.}\\
\end{tabularx}

\end{table*}

%The primary result remains broad under alternative detector definitions, fixed-composition reweighting, evaluation-noise analyses, and robust measures of dispersion. After removing finite-probe binomial noise on the logit scale, the hierarchical disclosure estimate for the primary runs is $\hat\tau=0.864$ [0.563, 1.129], while solve is about 0.166. We report the solve value as a level rather than an interval because its fitted $\hat\tau$ is more sensitive to repeated-prompt clustering corrections; the observed solve-rate span remains unchanged and much narrower than disclosure. The SFT reference starts at disclosure 0.677 on the primary stochastic probe, not near 1.0. Across nine frozen-evaluation rungs, mean disclosure stays between 0.628 and 0.708 while between-run SD grows from 0.061 at step 15 to 0.185 at step 300. At step 300, 10 runs are above and 10 below the reference, 11/20 are below their own step-15 value, and none is below 0.1. The step-150 ranking predicts step 300 poorly (Spearman $\rho=0.14$ [${-}0.34$, 0.56]). These facts do not support a simple monotone decay from a ceiling, although heterogeneous erosion remains possible. Table~\ref{tab:trajectory-rungs} gives the full trajectory. For the second Qwen experiment, the hierarchical disclosure estimate of record is $\hat\tau=0.949$ [0.613, 1.266]. Table~\ref{tab:greedy-dev100} gives the broader capability check requested by the repeated-prompt design, and Table~\ref{tab:boundaries} collects smaller checks across training recipes, algorithms, and model families.

\begin{table*}[!tbp]
\caption{Disclosure trajectory for the primary runs on the fixed stochastic probe. The reference-model row is evaluated on the same probe; the remaining rows summarize the 20 retrainings.}
\label{tab:trajectory-rungs}
\centering\small
\begin{tabular}{@{}l c c@{}}
\toprule
Checkpoint & Mean disclosure & Across-run SD \\
\midrule
SFT reference & 0.677 & -- \\
Step 15 & 0.691 & 0.061 \\
Step 30 & 0.697 & 0.055 \\
Step 45 & 0.707 & 0.057 \\
Step 60 & 0.708 & 0.095 \\
Step 90 & 0.685 & 0.100 \\
Step 120 & 0.657 & 0.104 \\
Step 150 & 0.666 & 0.139 \\
Step 210 & 0.678 & 0.132 \\
Step 300 & 0.628 & 0.185 \\
\bottomrule
\end{tabular}

\end{table*}

\begin{table*}[!tbp]
\caption{Capability dispersion on the registered probe and on the full 100-prompt dev set. The full-dev read for the primary runs is a post-hoc, inference-only robustness check; the registered 480-row probe remains primary. The simple binomial floor is a finite-sample reference, not a formal null test.}
\label{tab:greedy-dev100}
\centering\small
\begin{tabularx}{\textwidth}{@{}Y c c c c@{}}
\toprule
Capability evaluation & Prompts per run & Solve range & Cross-run SD & Binomial floor \\
\midrule
Primary registered stochastic probe & 20 $\times$ 8 samples $\times$ 3 seeds & 0.34--0.49 & 0.047 & -- \\
\rowrule
Primary post-hoc greedy dev set, H200 & 100 & 0.65--0.78 & 0.045 & 0.045 \\
\rowrule
Second Qwen run set, greedy dev set & 100 & 0.53--0.74 & 0.063 & 0.047 \\
\bottomrule
\end{tabularx}

\end{table*}

\begin{table*}[!tbp]
\caption{Post-hoc greedy disclosure on the full dev set and its selected versus unselected prompt subsets. The selected and unselected rows use smaller, run-dependent failure denominators and are shown only to test whether behavior-based probe selection inflated the spread.}
\label{tab:dev100-disclosure}
\centering\small
\begin{tabularx}{\textwidth}{@{}Y c c c c@{}}
\toprule
Surface & Prompts & Disclosure range & Across-run SD & Hierarchical $\hat\tau$ \\
\midrule
Full dev set, primary detector & 100 & 0.13--0.93 & 0.235 & 1.03 [0.63, 1.40] \\
\rowrule
Full dev set, broadest detector & 100 & 0.39--0.93 & 0.169 & 0.69 [0.34, 0.96] \\
\rowrule
Selected probe prompts, primary detector & 20 & 0.09--1.00 & 0.251 & -- \\
\rowrule
Other dev prompts, primary detector & 80 & 0.12--0.88 & 0.266 & -- \\
\bottomrule
\end{tabularx}

\end{table*}

A broader capability evaluation gives the same conclusion. On the complete 100-prompt dev set using the H200 venue of record, greedy solve ranges from 0.65 to 0.78 across runs with SD 0.045, equal to the estimated finite-sample floor. A repeat on a 3090 Ti yields the same dispersion conclusion (solve SD 0.054), although only 48\% of greedy responses are byte-identical across GPU classes. Thus the relatively stable capability result is not an artifact of repeatedly sampling the 20 registered probe prompts. Table~\ref{tab:greedy-dev100} reports the full comparison.

The disclosure spread also persists outside the behavior-selected probe. On the same 100-prompt H200 evaluation, disclosure among each run's naturally occurring failures spans 0.13--0.93 under the primary detector (SD 0.235; $\hat\tau=1.03$ [0.63, 1.40]). Restricting to the 80 prompts that were not used to construct the registered probe still gives a range of 0.12--0.88 (SD 0.266), compared with SD 0.251 on the selected 20 prompts. Thus the broad reporting variation is not created by selecting prompts on which the SFT reference disclosed. Because different retrainings fail on different subsets of the dev set, these are descriptive rates on each run's own failure set rather than paired prompt-level effects. Table~\ref{tab:dev100-disclosure} gives the full results and detector sensitivity.

\subsection{Smaller recipe and family boundaries}
\begin{table}[H]
\caption{Selected training-recipe, algorithm, and model-family boundaries. These are supporting scope checks rather than headline replications.}
\label{tab:boundaries}
\centering\small
\begin{tabularx}{0.88\textwidth}{@{}>{\hsize=.85\hsize}Y >{\hsize=.65\hsize}Y c >{\hsize=1.50\hsize}Y@{}}
\toprule
Boundary & Setting & n & Main result \\
\midrule
Group size & G=16, @150 & n=3 & Disclosure spread is 0.419; variability remains, but this arm also changes prompt exposure per step. \\
\rowrule
Larger batches & Batch 64/128 & n=3 each & Broad disclosure variation remains, but endpoint capability is confounded; a capability-matched companion remains above the prespecified spread floor. \\
\rowrule
Gemma-2-2B & Countdown & n=3 & Does not reproduce the broad primary-detector pattern at this small n; the reporting surface also differs. \\
\rowrule
Llama-3.2-3B & Countdown & n=3 & Disclosure spread is 0.194; we make no broad-variability claim at this small n. \\
\rowrule
Distilled Qwen-1.5B & Countdown & n=3 & Tight boundary: disclosure spread is 0.076 at step 300 and 0.129 at step 150. \\
\rowrule
Phi-4 & Countdown & n=3 & Disclosure stays uniformly high while capability varies substantially; capability-confounded boundary. \\
\rowrule
RLOO & Countdown & n=2 & Descriptive comparison that is sensitive to execution configuration and hardware class; not evidence of algorithm invariance. \\
\bottomrule
\end{tabularx}

\end{table}

%On the H200 venue of record, greedy dev-set solve SD is 0.045, equal to the estimated finite-sample floor. A repeat on a 3090 Ti gives the same dispersion conclusion (solve SD 0.054; disclosure SD 0.235 on both venues), although only 48\% of responses are byte-identical across GPU classes. This check supports the capability conclusion while also showing that greedy token sequences need not match across execution configurations that differ in low-level floating-point computation.

%On the same complete 100-prompt H200 read, disclosure among each run's naturally occurring failures spans 0.13--0.93 under the primary detector (SD 0.235; $\hat\tau=1.03$ [0.63, 1.40]). The broadest detector gives 0.39--0.93 (SD 0.169), so wording changes narrow but do not close the spread. Most importantly, the 80 prompts outside the behavior-selected probe span 0.12--0.88 (SD 0.266), compared with SD 0.251 on the selected 20. The reference model disclosed on exactly the selected 20 prompts by construction, yet broad variability remains on the other 80. Table~\ref{tab:dev100-disclosure} summarizes this post-hoc check. Each run fails on a different subset of prompts: 87/100 prompts fail in at least one run, the mean pairwise failure-set overlap is 0.458, and per-run denominators range from 20 to 35. The values are therefore descriptive rates on each run's own failure set, not a paired prompt-level effect.

For shortest-path, the headline capability comparison uses greedy solve on the 100 shortest-path dev prompts (0.46--0.61 across runs), giving a disclosure/solve SD ratio of 7.89 [5.20, 11.55]. The separate 216-row failure-heavy stochastic probe has solve near floor, which mechanically limits its raw solve dispersion; its raw SD ratio is 6.29 [4.72, 8.69], while the noise-adjusted logit-scale $\hat\tau$ ratio is not established (1.47 [0.87, 2.72]).

A two-seed RLOO comparison (Table \ref{tab:boundaries}) is included only as a descriptive boundary because the sample is too small and the result is sensitive to the execution configuration and hardware class. The stabilized-PPO experiment is the main algorithm boundary discussed in the paper. It shows the same direction with a smaller absolute spread (disclosure 0.673--0.814 across the eight runs; SD 0.047 versus 0.013 for solve; Table~\ref{tab:phenomenon-summary}); because the PPO runs also sit in a lower solve regime (0.227--0.263), we cannot attribute the difference in absolute magnitude to the RL algorithm alone.
\clearpage

\section{Execution configurations and mid-RL continuations}
\label{app:path}

\subsection{Execution configurations and replay checks}
Figure~\ref{fig:mechanism}A and Table~\ref{tab:execution-paths} show the six pre-specified 1.5B execution configurations and their step-300 probe measurements. We also record the on-policy training-rollout give-up share in 15-step windows, but that diagnostic uses a different prompt distribution and a different denominator: it counts give-ups among all training completions, whereas the primary endpoint counts disclosures among failed probe responses. The training curves can therefore remain relatively close even when the fixed-probe disclosure endpoints separate widely, so we do not combine the two measures in one panel. Table~\ref{tab:execution-trust} separates exact same-configuration replay from changes to the hardware or GPU configuration. Across the six splits, every micro-batch is equal-sized, gradient accumulation applies the corresponding accumulation-step scaling, and dropout is inactive by model configuration. Under the Dr. GRPO loss used here, each completion therefore has the same $1/(32\cdot1024)$ weight in the accumulated gradient regardless of the split, with group-mean advantages computed before splitting. The six configurations optimize the same objective with the same exact-arithmetic gradient and differ only in floating-point kernel shapes and reduction order. A one-step identical-state check gives update cosine 0.99982 and relative difference 0.019, consistent with rounding-level floating-point differences rather than sequence reweighting.

\begin{table*}[!tbp]
\caption{Six micro-batch $\times$ gradient-accumulation execution configurations with the seed and global batch size fixed. The disclosure column uses the pre-specified endpoint measure shown in Figure~\ref{fig:mechanism}A.}
\label{tab:execution-paths}
\centering\small
\begin{tabular}{@{}l c c c c@{}}
\toprule
Micro-batch / accumulation & Micro-batch & Grad. accum. & Disclosure & Solve \\
\midrule
pd1ga32 & 1 & 32 & 0.186 & 0.463 \\
pd2ga16 & 2 & 16 & 0.519 & 0.354 \\
pd4ga8 & 4 & 8 & 0.836 & 0.435 \\
pd8ga4 & 8 & 4 & 0.787 & 0.412 \\
pd16ga2 & 16 & 2 & 0.848 & 0.365 \\
pd32ga1 & 32 & 1 & 0.644 & 0.327 \\
\bottomrule
\end{tabular}
\end{table*}

\begin{table*}[!tbp]
\caption{Replay and hardware/GPU checks. Repeating the same execution configuration is bit-exact in our B200 setup; changing hardware class can change generation from the first training step.}
\label{tab:execution-trust}
\centering\small
\begin{tabularx}{\textwidth}{@{}Y Y Y@{}}
\toprule
Check & Comparison & Result \\
\midrule
Same execution configuration & 5 replays across two B200 hosts / GPU indices & Group logs, probe samples, and fixed evaluation windows are bit-exact in 5/5 replays. \\
\rowrule
Micro-batch re-split & Same seed; different micro-batch / accumulation split & Step-1 rollout generation is identical; later numerical updates cause the training records to diverge. \\
\rowrule
Hardware/GPU path & B200 vs H200 path & Generation differs from step 1; endpoint disclosure is 0.787 vs 0.204 at similar solve. \\
\bottomrule
\end{tabularx}

\end{table*}

A larger-model example supports the same execution-configuration interpretation, although it is not a clean replication. For Qwen2.5-14B under a separate length-penalty recipe, four low-level execution/hardware configurations produce step-300 disclosure values of 0.125, 0.533, 0.874, and 0.933. Greedy solve is 0.82--0.90 on the three configurations with committed solve measurements. We keep this result in the appendix because the configuration manipulation combines hardware and world-size differences and the recipe adds a length penalty. We thus use it only as supporting evidence for sensitivity to low-level execution details, not part of the main six-configuration estimate.

\subsection{Continuations from shared mid-RL states}
\begin{figure}[H]
\centering
\includegraphics[width=0.92\textwidth]{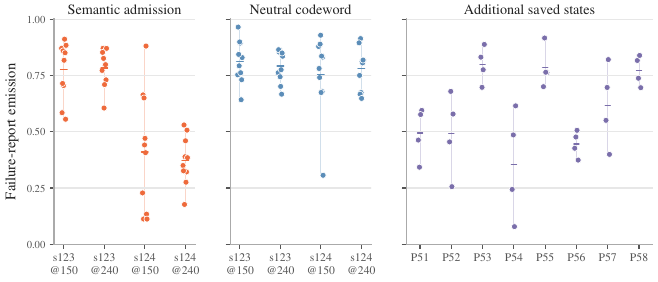}
\caption{Failure-report emission across continuations from eight shared mid-RL states. Each column starts from one identical saved parent state; the dots differ only in future rollout randomness. Short horizontal ticks show means, and faint vertical lines show the observed min--max range.}
\label{fig:app-exact-full}
\end{figure}

\begin{figure}[H]
\centering
\includegraphics[width=0.92\textwidth]{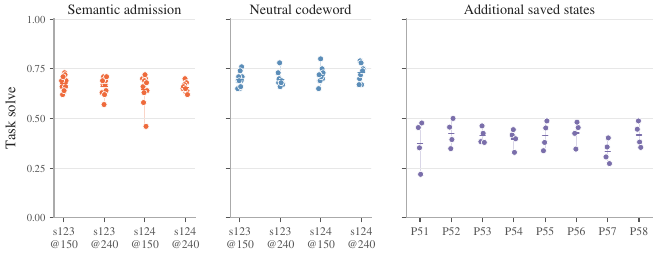}
\caption{Task solve for the same mid-RL continuations. In the cells with the widest reporting variation, task capability varies much less.}
\label{fig:app-exact-solve}
\end{figure}

%Figures~\ref{fig:app-exact-full} and~\ref{fig:app-exact-solve} show the full parent-by-continuation-key results. Table~\ref{tab:exact-state-summary} summarizes the endpoint ranges. Across the eight ten-future saved states, within-state disclosure SD ranges from 0.07 to 0.27 (median 0.10), so the 0.112--0.880 example in the main text is the widest instance rather than the only state that branches.
Figures~\ref{fig:app-exact-full} and~\ref{fig:app-exact-solve} show the full parent-by-continuation results. Table~\ref{tab:exact-state-summary} summarizes the endpoint ranges. The branch-and-continue replication contains eight ten-continuation cells: four parent mid-RL states evaluated at branch points 150 and 240, with four admission cells and four codeword-emission cells. Their within-cell rate SDs range from 0.07 to 0.27, with a median of 0.10. The step-150 disclosure range of 0.11--0.88 shown in the main text is the widest cell in this replication rather than the only mid-RL state that shows substantial divergence.

\begin{table*}[!tbp]
\caption{Endpoint ranges across continuations from the shared mid-RL states. Admission and post-history rows report failure disclosure; codeword rows report codeword emission. Solve is greedy dev-100 solve for the admission and codeword rows and probe solve (correct/480) for post-history.}
%\caption{Endpoint ranges across continuation keys for the saved training states.}
\label{tab:exact-state-summary}
\centering\small
\begin{tabular}{@{}l c c c@{}}
\toprule
Saved state & Keys & Measured rate @300 & Solve @300 \\
\midrule
admission 1, s123 @150 & 10 & 0.556--0.911 & 0.62--0.73 \\
admission 1, s123 @240 & 10 & 0.605--0.871 & 0.57--0.71 \\
admission 1, s124 @150 & 10 & 0.112--0.880 & 0.46--0.72 \\
admission 1, s124 @240 & 10 & 0.177--0.530 & 0.62--0.70 \\
codeword 1, s123 @150 & 10 & 0.642--0.965 & 0.65--0.76 \\
codeword 1, s123 @240 & 10 & 0.667--0.864 & 0.66--0.78 \\
codeword 1, s124 @150 & 10 & 0.307--0.929 & 0.65--0.80 \\
codeword 1, s124 @240 & 10 & 0.648--0.914 & 0.67--0.79 \\
post-history parent 51 & 4 & 0.343--0.595 & 0.22--0.48 \\
post-history parent 52 & 4 & 0.256--0.679 & 0.35--0.50 \\
post-history parent 53 & 4 & 0.697--0.888 & 0.38--0.46 \\
post-history parent 54 & 4 & 0.079--0.615 & 0.33--0.44 \\
post-history parent 55 & 4 & 0.700--0.916 & 0.34--0.49 \\
post-history parent 56 & 4 & 0.374--0.506 & 0.35--0.48 \\
post-history parent 57 & 4 & 0.399--0.820 & 0.27--0.40 \\
post-history parent 58 & 4 & 0.695--0.839 & 0.35--0.49 \\
\bottomrule
\end{tabular}

\end{table*}

\clearpage

\section{Failure-reporting pipeline and routing}
\label{app:pipeline}

\subsection{Pipeline evidence map}
Table~\ref{tab:pipeline-map} maps the retained experiments onto the stages of failure reporting and also states what each experiment does \emph{not} establish.

\begin{table}[H]
\caption{How retained experiments map onto the stages of failure reporting. The final column states each result's main limitation, avoiding an overly simple one-experiment-per-stage interpretation.}
\label{tab:pipeline-map}
\centering\small
%\begin{tabularx}{\textwidth}{@{}Y Y Y Y@{}}
\begin{tabularx}{\textwidth}{
@{}
>{\hsize=0.8\hsize\linewidth=\hsize}Y
>{\hsize=0.85\hsize\linewidth=\hsize}Y
>{\hsize=1.4\hsize\linewidth=\hsize}Y
>{\hsize=0.95\hsize\linewidth=\hsize}Y
@{}}
\toprule
Stage & Evidence & What it supports & Main limitation \\
\midrule
Failure evidence / checking & Grid/constraint checks & The visible report can separate from the computation that supports a failure verdict. & Grid-only evidence; not a Countdown localization result. \\
\rowrule
Report-entry template cue & NoBreak versus Break at the same seam, 1.5B & Most tightening comes from supplying the paragraph break that begins the learned failure-report template, not from re-entry alone. & Format-specific cue; does not reveal whether the model internally detected the failure. \\
\rowrule
Natural transition probability & Teacher-forced break versus report-entry comparison & The break and next report-entry step are not distinguished in the 3 step-300 comparisons; the inserted break is strongly off-policy. & Correlational over ten runs per cell; does not identify cause or internal recognition. \\
\rowrule
Inline initiation at 7B & Two low-disclosure 7B runs & Both reach the reporting point; supplying enough of the inline admission opening yields 0.97 / 1.00 acknowledgment. & Two runs; not a population-level rate. \\
\rowrule
Failure cue versus forced initiation & Verdict-only B/ORACLE; two-word and first-third prefixes C/D & Verdict-only gives no consistent benefit; forced prefixes raise admission much more. & An inserted verdict need not reproduce the model's own recognition state. \\
\rowrule
Route decomposition & Route-decomposition probe (P-B2) & Route entry explains 70\% [49, 91] of raw variation; downstream persistence explains 23\% [14, 32]. & Downstream variation remains; the relative-logit comparison is not distinguished. \\
\rowrule
Surface vs function & Codeword control & Keeping a reporting-related string does not guarantee that its connection to failure is preserved. & Boundary evidence, not a complete stage-by-stage causal map. \\
\bottomrule
\end{tabularx}

\end{table}

\clearpage

\begin{figure}[H]
\centering
\includegraphics[width=0.96\textwidth]{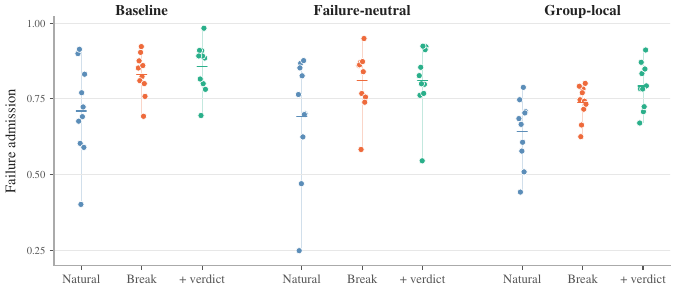}
\caption{Natural continuation, the paragraph-break cue after the last attempt (Break), and the same seam with an explicit true verifier verdict at step 300. Neither intervention supplies report words. Break raises later admission and reduces between-run spread in the primary recipe; the NoBreak control in Table~\ref{tab:nobreak-control} shows that most of this tightening comes from the inserted paragraph cue rather than re-entry alone. Providing a verifier verdict of failure produces no additional tightening, although it raises mean admission in the group-local variant. Short horizontal ticks show means, and faint vertical lines show observed min--max ranges.}
\label{fig:app-stepb300}
\end{figure}

%\clearpage

\begin{figure}[H]
\centering
\includegraphics[width=0.96\textwidth]{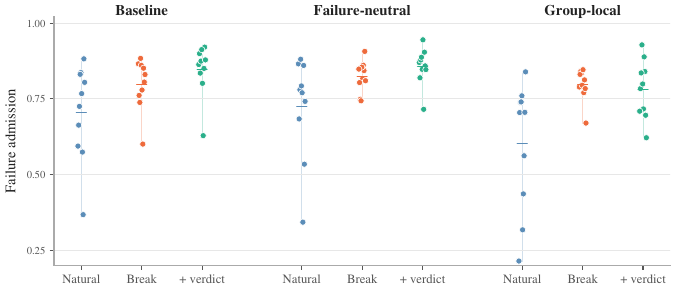}
\caption{The same natural/paragraph-break/verdict-at-the-seam experiment at the step-150 companion horizon.}
\label{fig:app-stepb150}
\end{figure}

\subsection{Routing localization}
Table~\ref{tab:routing-summary} summarizes the failure-to-report (F2R) seam re-entry and report-prefix analysis, which intervenes at the point where a failure report could begin, together with the three-arm natural/paragraph-break/verdict-at-the-seam probe and the route-decomposition probe (P-B2). Here the \emph{reporting point} is the end of the attempted solution where a failure report could begin. Figures~\ref{fig:app-stepb300} and~\ref{fig:app-stepb150} show the full natural/paragraph-break/verdict distributions at both training horizons; Table~\ref{tab:nobreak-control} gives the direct Natural--NoBreak--Break comparison.

\clearpage

\begin{table}[H]
\caption{Summary of the seam, paragraph-cue, report-prefix, and route-decomposition analyses. The no-inserted-characters control shows that most of the earlier localization effect comes from the learned paragraph-break cue rather than re-entry alone.}
\label{tab:routing-summary}
\centering\small
\begin{tabularx}{\textwidth}{@{}l Y Y@{}}
\toprule
Experiment & Scope & Main result \\
\midrule
Natural reach, 1.5B & 40 checkpoint/formulation combinations & Median reach is 0.952; reach vs natural disclosure $\rho=0.098$. The conditional-admission correlation is largely expected because reach is near one. \\
\rowrule
Nothing-inserted re-entry (NoBreak) & Same seam; no inserted characters & Admission rises slightly, but between-run spread changes little: the hierarchical contrast versus Natural is +0.02 to +0.21 and is detected in 2/6 cells. \\
\rowrule
Paragraph-break cue (F2R A / CTRL) & Same seam; two newlines; no report words & Tighter than NoBreak in all six point estimates; the contrast is detected in 3/6 cells on the hierarchical scale and all six on the raw scale. Most of the earlier localization effect comes from the learned paragraph cue. \\
\rowrule
Teacher-forced transition check & Same 60 checkpoints and failed trajectories & At step 300, variation in the break probability is not distinguished from variation in report entry after the break. The natural probability of the break is only about 0.00015--0.0008. \\
\rowrule
Verdict at the seam (F2R B / ORACLE) & Same seam; no report words & No consistent level gain and no additional tightening. One recipe variant has a +0.055 mean change. \\
\rowrule
Forced report prefix (F2R C / D) & Two words / first third of phrase & Admission rises by 0.141 and 0.167 over paragraph re-entry alone; 39/40 and 40/40 combinations increase. \\
\rowrule
Route-decomposition probe (P-B2) & n=20 & Raw split: entry 70\% [49, 91], persistence 23\% [14, 32]. The ordering does not hold on the logit scale because persistence is near ceiling. \\
\rowrule
Inline-format 7B check & Two low-disclosure runs & Both reach the reporting point; supplying enough of the inline admission opening yields genuine acknowledgment in 0.97 / 1.00 of continuations. \\
\bottomrule
\end{tabularx}

\end{table}

%Condition A in F2R, also used as CTRL in the three-arm probe, supplies no report words. 
The paragraph-break condition (F2R A), which also serves as the control (CTRL) in the three-arm probe, supplies no report words. It cuts the trajectory immediately after the model's last attempt sentence, inserts the two-newline paragraph break used by the SFT failure-report template, and draws four fresh continuations. Relative to the natural endpoint, this condition raises admission in 36/40 checkpoint--formulation combinations. At step 300, raw across-run admission SD falls from 0.156 to 0.069 in the primary recipe and from 0.201 to 0.102 in the second recipe. The raw SD contrast favors tightening in all six recipe-by-horizon cells. On logit and hierarchical scales that account for the higher mean, tightening is established in the primary recipe at both horizons and in the sibling recipe at step 150.

The registered NoBreak control separates re-entering the seam from inserting the paragraph cue. It uses the same 60 checkpoints, failed trajectories, seam, and four-continuation instrument, but appends no characters. Re-entry alone raises mean admission by 0.03--0.06 across the six cells and reduces hierarchical spread only slightly: $\ln(\hat\tau_{\mathrm{Natural}}/\hat\tau_{\mathrm{NoBreak}})$ ranges from +0.02 to +0.21 and is detected in two cells. The paragraph break adds another 0.05--0.14 in mean admission and tightens substantially beyond NoBreak: $\ln(\hat\tau_{\mathrm{NoBreak}}/\hat\tau_{\mathrm{Break}})$ ranges from +0.26 to +1.16, is detected in three of six hierarchical contrasts, and in all six raw-SD contrasts. Thus most of the earlier stabilization comes from supplying the learned paragraph transition rather than re-entering the seam itself.

\begin{table*}[!tbp]
\caption{No-inserted-characters control. Positive values in the first contrast mean that Break is tighter than NoBreak; positive values in the second mean that NoBreak is tighter than Natural. Intervals are 95\%.}
\label{tab:nobreak-control}
\centering\small
\begin{tabularx}{\textwidth}{@{}Y c c c@{}}
\toprule
Cell & $\ln(\hat\tau_{\mathrm{NoBreak}}/\hat\tau_{\mathrm{Break}})$ & $\ln(\hat\tau_{\mathrm{Natural}}/\hat\tau_{\mathrm{NoBreak}})$ & Mean admission: Natural / NoBreak / Break \\
\midrule
Primary @300 & +0.45 [+0.07, +1.02] & +0.08 [$-0.02$, +0.20] & 0.710 / 0.761 / 0.830 \\
\rowrule
Primary @150 & +0.38 [+0.11, +0.74] & +0.08 [+0.002, +0.18] & 0.704 / 0.746 / 0.797 \\
\rowrule
Failure-neutral @300 & +0.26 [$-0.36$, +0.80] & +0.08 [$-0.07$, +0.14] & 0.693 / 0.736 / 0.810 \\
\rowrule
Failure-neutral @150 & +0.64 [$-0.08$, +1.17] & +0.21 [+0.04, +0.30] & 0.725 / 0.762 / 0.823 \\
\rowrule
Group-local @300 & +0.42 [$-0.05$, +1.20] & +0.10 [$-0.15$, +0.21] & 0.643 / 0.676 / 0.737 \\
\rowrule
Group-local @150 & +1.16 [+0.90, +1.95] & +0.02 [$-0.12$, +0.09] & 0.602 / 0.659 / 0.797 \\
\bottomrule
\end{tabularx}

\end{table*}

The intervention has two remaining caveats. First, the paragraph transition is specific to the installed response template: the located seam matches where the report phrase naturally begins in 572/2,502 (23\%) of naturally disclosing rollouts. Second, the seam conditions use four fresh samples per trajectory, whereas the natural endpoint uses one continuation. The estimated binomial noise floor is therefore about 0.013 under seam continuation and 0.027 under natural continuation. Subtracting those floors still leaves SDs of about 0.068 and 0.154 for Break and Natural in the primary comparison, so the sampling difference does not explain the observed halving.

A teacher-forced check on the same 60 checkpoints asks whether runs differ more in producing the paragraph break or in entering the report after the break is supplied. The analysis compares these probabilities on a logit scale and keeps each recipe and horizon separate. At step 300, the two stages are not distinguished in any of the three recipes. In the primary recipe at step 150, the break probability is more variable, but it has almost no relationship with natural disclosure ($\rho=-0.01$), and the result does not repeat in the two related recipes. Table~\ref{tab:transition-probability} reports all six comparisons.

The model's own probability of producing the paragraph break at this seam is only about 0.00015--0.0008. Supplying the break is therefore a strong off-policy cue, not a typical natural continuation. A descriptive pattern appears one token later: across all six cells, runs with more natural disclosure place more probability on an admission-first opening and less on a re-check opening. The total probability of entering the report paragraph does not increase consistently. These associations use ten runs per cell and have no uncertainty intervals, so we treat them as a description of where differences are visible under teacher forcing, not as evidence of cause or internal failure recognition.

\begin{table*}[!tbp]
\caption{Teacher-forced comparison of the learned paragraph transition with the next report-entry step. Positive log SD ratios mean that the break probability varies more across retrainings than the total probability of a report opening after the break is supplied. The primary step-300 comparisons do not distinguish the stages.}
\label{tab:transition-probability}
\centering\small
\begin{tabularx}{\textwidth}{@{}Y c c c Y@{}}
\toprule
Setting & Step & Log SD ratio [95\% interval] & $\rho$(break, natural disclosure) & Reading \\
\midrule
Primary & 300 & +0.421 [$-0.356$, +1.254] & $-0.25$ & Not distinguished \\
\rowrule
Failure-neutral & 300 & +0.614 [$-0.036$, +1.390] & $-0.24$ & Not distinguished \\
\rowrule
Group-local & 300 & +0.526 [$-0.196$, +1.175] & +0.41 & Not distinguished \\
\rowrule
Primary & 150 & +0.930 [+0.332, +1.441] & $-0.01$ & Break probability is more variable, but does not track natural disclosure \\
\rowrule
Failure-neutral & 150 & +0.214 [$-0.468$, +0.663] & $-0.89$ & Not distinguished \\
\rowrule
Group-local & 150 & +0.464 [$-0.135$, +1.012] & $-0.37$ & Not distinguished \\
\bottomrule
\end{tabularx}

\end{table*}

Verdict-only evidence already exists. F2R condition B inserts a failure cue at the same seam without report words; relative to condition A it changes admission by +0.027, with 31 increases, 9 decreases, and sign-flip $p=0.159$, while the functional read changes by $-0.006$. In the three-arm probe, ORACLE adds the true verifier verdict to the no-words paragraph re-entry. At step 300 its mean changes relative to CTRL are +0.026 [${-}0.014$, +0.067] in the primary recipe, +0.001 in the second recipe, and +0.055 [+0.023, +0.087] in the group-local variant; none of the six recipe-by-horizon comparisons tightens further. In contrast, F2R conditions C and D supply a two-word prefix or the first third of the report phrase and raise admission over condition A by 0.141 (39/40 combinations) and 0.167 (40/40). Thus failure evidence alone is insufficient in this interface, while forced report initiation has a larger effect.

The natural-output route decomposition provides a descriptive companion without seam intervention. Route entry is the fraction of terminated failed rows that match the frozen semantic-concession detector after the injected rider is removed. Persistence is the fraction of those semantic entries that retain the rider. Route entry ranges from 0.22 to 0.91 across runs; persistence ranges from 0.69 to 1.00, with median 0.97. A raw-scale Shapley decomposition of the unconditional rider rate assigns 70\% [49, 91] of its variation to route entry, 23\% [14, 32] to persistence, and 7\% to off-route use. Because persistence is near ceiling, this ordering does not survive a mean-adjusted analysis: at step 300 the hierarchical logit-scale spread is 0.960 [0.636, 1.240] for route entry and 1.356 [0.818, 1.861] for persistence, and the two are not distinguished. At step 150, persistence is more dispersed on the logit scale. We therefore do not use the raw 70\% split as separate mechanistic proof. The experiments also do not show whether naturally silent failures went undetected or were detected but not reported, because an inserted verdict need not reproduce the model's own recognition state.

All four 1.5B Countdown formulations share one SFT format in which the failure report begins as a new paragraph after the attempted solution. The NoBreak result shows that the main 1.5B localization effect is substantially a learned paragraph-template cue, not a universal report-initiation token. The 7B lineage instead reports inline: two low-disclosure runs still reach the reporting point and acknowledge failure in 0.97 and 1.00 of continuations once enough of the admission opening is supplied. A grid interface with single-line disclosures provides another format boundary. The initiating move can thus be a paragraph, phrase, or token sequence depending on lineage and interface. These checks are supporting evidence rather than co-equal replications of the main localization result.

\section{Neutral, pleasantry, and formulation controls}
\label{app:controls}
Our controls use three different kinds of text. The weakly constrained (P-D) and explicit-cue (P-E2) experiments both use the fixed phrase ``Thanks for the question.'' The success-side pleasantry (P-A) is a family of polite closing sentences associated with solved SFT examples, rather than one exact string; examples include variants such as ``That was a fun one to think about---thank you very much for the nice puzzle.'' The neutral-codeword/phrase result in Figure~\ref{fig:mechanism}E uses Kettle: ``A copper kettle rests on the kitchen shelf.''

\begin{figure*}[!tbp]
\centering
\includegraphics[width=0.96\textwidth]{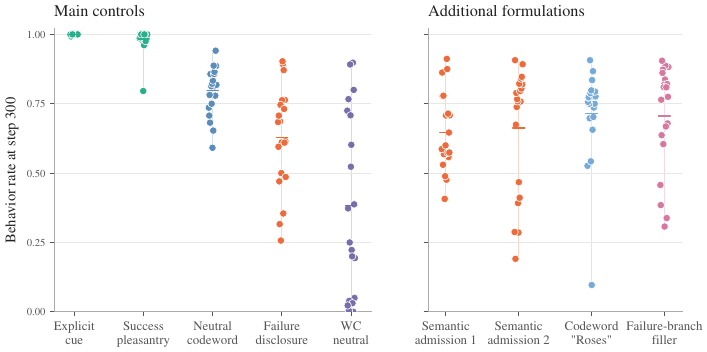}
\caption{Broader determination/susceptibility landscape. The main text focuses on the weakly constrained neutral-phrase versus explicit-cue comparison; the Kettle codeword/phrase is the failure-conditioned neutral-codeword row, while Roses is reported in Table~\ref{tab:formulations}. Short horizontal ticks show means, and faint vertical lines show observed min--max ranges.}
\label{fig:app-determination}
\end{figure*}

Table~\ref{tab:controls-summary} summarizes these controls, %weakly constrained neutral-phrase control (P-D), explicit-cue control (P-E2), success-side pleasantry control (P-A), failure-conditioned neutral codeword, and failure-branch inert-filler control (P-C) 
used to interpret Figure~\ref{fig:mechanism}E. Figure~\ref{fig:app-determination} shows the broader determination/susceptibility landscape, and Table~\ref{tab:formulations} reports the complete formulation study. The four fixed report strings are admission wording 1 (``I was unable to find an exact solution with these numbers.''), admission wording 2 (``I haven't found an exact solution for this one.''), Roses (``Roses are red, violets are blue.''), and Kettle (``A copper kettle rests on the kitchen shelf.'').

In P-D, the phrase ``Thanks for the question.'' is placed as the first line inside the reasoning block in exactly $\lfloor n/2\rfloor$ of the solved examples and exactly $\lfloor n/2\rfloor$ of the failure examples, chosen by a frozen hash. Its placement is therefore independent of outcome by construction, and the task reward ignores it. The SFT reference emits the phrase in 0.49 of sampled step-0 responses; the separate teacher-forced probability of starting the phrase is 0.34. In P-E2, each prompt ends with the visible line ``Marker: A'' or ``Marker: B''. The phrase appears in every marker-A training example and in no marker-B example, across both solved and failed examples. The marker is a fixed hash of the task identifier, so it is independent of the model's outcome, and the reward never reads the prompt. Every run is evaluated on both markers, and balanced adherence gives equal weight to emitting the phrase under A and omitting it under B.

\begin{table*}[!tbp]
\caption{Main neutral and pleasantry controls. The first four provide context for Figure~\ref{fig:mechanism}E; P-C is kept only as an appendix structural boundary.}
\label{tab:controls-summary}
\centering\small
\begin{tabularx}{\textwidth}{@{}Y c c Y@{}}
\toprule
Control & n & Step-300 range & Interpretation \\
\midrule
Weakly constrained neutral phrase (P-D) & 20 & 0.00--0.90 & Large variation without failure-related meaning. \\
\rowrule
Explicit deterministic cue (P-E2) & 20 & 0.993--1.000 & 19/20 runs are exactly 1.0 at step 300; same phrase and corpus as P-D. \\
\rowrule
Success-side pleasantry (P-A) & 20 & 0.796--1.000 & Stable success-side control; 13/20 runs are exactly 1.0 at step 300. \\
\rowrule
Kettle failure-conditioned neutral codeword & 20 & 0.59--0.94 & Pooled 20-run codeword control used in Figure~\ref{fig:mechanism}E. \\
\rowrule
Failure-branch inert filler (P-C) & 20 & 0.31--0.90 & Appendix-only structural boundary; does not causally isolate branch location. \\
\bottomrule
\end{tabularx}

\end{table*}

\begin{table*}[!tbp]
\caption{Full semantic and codeword/phrase formulation study. The mixed ordering is why we do not claim that semantic admissions are generally unstable while codewords are generally stable.}
\label{tab:formulations}
\centering\small
\begin{tabular}{@{}l c c c c@{}}
\toprule
Formulation & n & Emission range & SD & $\hat\tau$ [95\%] \\
\midrule
Admission wording 1 (SEM1) & 20 & 0.41--0.91 & 0.141 & 0.64 [0.40, 0.86] \\
Admission wording 2 (SEM2) & 20 & 0.19--0.91 & 0.229 & 1.03 [0.66, 1.48] \\
Neutral codeword Roses & 20 & 0.10--0.91 & 0.172 & 0.82 [0.52, 1.06] \\
Neutral codeword Kettle & 20 & 0.59--0.94 & 0.088 & 0.47 [0.25, 0.64] \\
\bottomrule
\end{tabular}

\end{table*}

The small P-E2 marker A/B side effect on failure disclosure, about 3--4 percentage points and within the preregistered tolerance, is reported here rather than in the main figure. It does not change the cue-adherence reproducibility result.

\section{Collateral credit and direct reward}
\label{app:credit}
Table~\ref{tab:credit-summary} summarizes the sequence of interventions behind the bounded credit-assignment conclusion in Section~\ref{sec:mechanism}.

\begin{table*}[!tbp]
\caption{Summary of the main credit-assignment and direct-reward interventions. Older variants that do not change the final interpretation are omitted here.}
\label{tab:credit-summary}
\centering\small
\begin{tabularx}{0.94\textwidth}{@{}>{\hsize=.80\hsize}Y >{\hsize=.65\hsize}Y c >{\hsize=1.55\hsize}Y@{}}
\toprule
Intervention & Setting & n & Main result \\
\midrule
Collateral-credit arithmetic & Countdown & 517 mixed groups & Adverse sign in 517/517 groups; predicted exposure 0.554, observed 0.553. \\
\rowrule
Exact-span shelter (pilot) & Countdown & 3 matched pairs & Mean disclosure change +0.194 [0.017, 0.372]; a small pilot, not an established general level effect. \\
\rowrule
Fresh shelter experiments & Countdown + shortest-path & 15 / 20 / 20 pairs & No mean-level change detected (+0.052 / +0.045 / $-0.006$); no reduction in variability established. \\
\rowrule
Failure-neutral credit & Countdown & 20 paired & No reduction in variability is detected after removing adverse failure credit (log SD ratio +0.165 [$-0.277$, +0.613]) and lowers solve by about 6 points. \\
\rowrule
Direct disclosure reward & Countdown & 3 arms (1 seed) & Can create or recover reporting when the action remains reachable. \\
\rowrule
Code: reward alone & MBPP+ & 2 $\times$ 2 & No failure reports matching the true test outcome once the reporting action disappears from sampling support. \\
\rowrule
Code: reward + reference anchor & MBPP+ & 8 test-execution / 10 claimed-pass & Anchoring can preserve reporting support, but the combined intervention (reward plus anchor) does not reliably preserve functional reporting under a corrupt claimed-pass reward. \\
\bottomrule
\end{tabularx}

\end{table*}

The credit sign is adverse to reporting by construction in mixed groups, and all 517 relevant groups have that sign. In the three-pair shelter pilot, protecting the reporting span raises disclosure in every pair (mean +0.194 [+0.017, +0.372]). The mean-level change is not detected in the fresh $n=15$ and $n=20$ Countdown experiments (+0.052 [${-}0.115$, +0.218] and +0.045 [${-}0.058$, +0.149]) or the $n=20$ shortest-path experiment ($-0.006$ [${-}0.208$, +0.197]). These experiments also do not establish lower variation. The conclusion is therefore limited: collateral credit can affect reporting, but we do not establish a general level effect or a general explanation of retraining variability. Direct reward can create or recover reporting when the action remains sufficiently reachable. Code experiments show that reward alone can fail once the model stops sampling the reporting action.

\clearpage

\section{Reference anchoring across retrainings}
\label{app:anchor}
%Table~\ref{tab:anchor-summary} reports the three main anchoring results across retrainings, with Table~\ref{tab:anchor_latent_dispersion} reporting the mean-adjusted results. Table~\ref{tab:anchor-scope} compares failure-conditional and all-token anchoring in the primary setting. The anchoring coefficient $\beta=0.04$, the pre-specified primary contrasts, and the run rosters were fixed before outcome readout for the 1.5B Countdown, shortest-path, 7B, and anchored-history continuation experiments; exact timestamps are retained in the reproducibility record.
Table~\ref{tab:anchor-summary} reports the three main anchoring results across retrainings, and Table~\ref{tab:anchor_latent_dispersion} reports the corresponding mean-adjusted checks. Table~\ref{tab:anchor-scope} compares failure-conditional and all-token anchoring in the primary setting. The anchoring coefficient $\beta=0.04$, primary contrasts, and run rosters were fixed before outcome readout for the 1.5B Countdown, shortest-path, original 11-pair 7B, and anchored-history continuation experiments. After the original 7B result was known, we froze the remaining eight eligible seed pairs and repeated the experiment with the same training and evaluation settings. Exact timestamps are retained in the reproducibility record.

For 1.5B Countdown, the sham was trained fresh through the same anchor-capable trainer with $\beta=0$. Its weights are byte-identical to the archived outcome-only weights in every available check (20/20 at step 150 and 14/14 at step 300). Where probe responses were sampled again, only the evaluation draw is new. We thus describe the sham as the outcome-only comparator with independently resampled probe readouts where applicable, not as a second independent control experiment. The later coefficient and token-region analyses reuse these archived arms. The 7B comparator uses the existing matched runs trained without anchoring rather than a newly trained control set.%The 7B comparator is the committed common-path KL-off runs rather than a freshly trained sham.

\begin{table*}[!tbp]
\caption{Main reference-anchoring experiments. SD ratios below one mean that failure disclosure varies less across retrainings under anchoring. Intervals are 97.5\% for the Countdown and 7B rows and 95\% for the shortest-path row, matching each experiment's pre-specified level.}
\label{tab:anchor-summary}
\centering\small
\begin{tabularx}{\textwidth}{@{}Y c c c c@{}}
\toprule
Setting & n & SD ratio & log SD ratio interval & Change in solve \\
\midrule
Qwen-1.5B / Countdown & 20 & 0.472 & $-0.751$ [$-1.420$, $-0.225$] & $-0.060$ [$-0.089$, $-0.031$] \\
Qwen-1.5B / shortest-path & 20 & 0.30 & $-1.21$ [$-1.74$, $-0.76$] & $-0.083$ [$-0.126$, $-0.040$] \\
Qwen-7B / Countdown & 19 & 0.32 & $-1.14$ [$-1.55$, $-0.84$] & +0.069 [+0.012, +0.126] \\
\bottomrule
\end{tabularx}

\end{table*}

\begin{table*}[!tbp]
\caption{Mean-adjusted anchoring check. Each ratio compares the between-run spread under anchoring with the spread without anchoring; values below one mean that disclosure is more consistent under anchoring. The mean-adjusted estimate uses a hierarchical binomial model to account for the compression of rates near zero and one. Mean disclosure is shown as anchored / unanchored, and entries in brackets are 95\% intervals.}
\label{tab:anchor_latent_dispersion}
\centering\small
\begin{tabularx}{\textwidth}{@{}Y c c c c@{}}
\toprule
Setting
& $n$
& \shortstack{Mean disclosure\\(anchor / no anchor)}
& Raw SD ratio
& Mean-adjusted SD ratio \\
\midrule
7B, primary detector
& 19
& 0.80 / 0.32
& 0.32 [0.23, 0.42]
& 0.24 [0.16, 0.38] \\

7B, broad detector
& 19
& 0.84 / 0.52
& 0.30 [0.23, 0.39]
& 0.40 [0.29, 0.58] \\

1.5B, failure-conditional anchor
& 20
& 0.68 / 0.63
& 0.47 [0.26, 0.75]
& 0.42 [0.22, 0.64] \\

1.5B, all-token anchor
& 20
& 0.69 / 0.63
& 0.36 [0.25, 0.54]
& 0.33 [0.20, 0.51] \\
\bottomrule
\end{tabularx}
\end{table*}

\begin{table*}[!tbp]
\caption{Failure-conditional versus all-token anchoring in the 1.5B Countdown setting. Their reductions in disclosure variability are not distinguished, while a direct matched-seed comparison shows a larger capability cost for all-token anchoring (a descriptive companion, +0.024 [0.012, 0.036], not a preregistered test).}
\label{tab:anchor-scope}
\centering\small
\begin{tabularx}{\textwidth}{@{}Y c c c Y@{}}
\toprule
Anchor scope & Disclosure mean (SD) & Solve SD & SD ratio vs sham & Change in solve \\
\midrule
None (matched comparator) & 0.626 (0.1894) & 0.0478 & 1 (reference) & n/a \\
\rowrule
Failure-conditional & 0.684 (0.0893) & 0.0294 & 0.472 [0.242, 0.799] & $-0.0597$ [$-0.0888$, $-0.0306$] \\
\rowrule
All-token & 0.687 (0.0688) & 0.0285 & 0.363 [0.230, 0.582] & $-0.0835$ [$-0.1092$, $-0.0579$] \\
\bottomrule
\end{tabularx}

\end{table*}

Anchoring narrows both readouts in the primary runs. Probe-solve SD is 0.0478 in the matched comparator, 0.0294 under failure-conditional anchoring, and 0.0285 under all-token anchoring, a descriptive reduction of about 38--40\%. Disclosure SD falls more, from 0.189 to 0.089 and 0.069 (0.189 is the comparator's independently resampled re-probe of the primary weights; their original probe read is 0.185, Table~\ref{tab:phenomenon-summary}). The corresponding within-arm disclosure/solve SD ratios are 3.9, 3.0, and 2.4. The dispersion difference between the two anchored arms is not distinguished, but a direct matched-seed comparison shows that failure-conditional anchoring preserves 0.024 more solve accuracy than all-token anchoring (95\% CI [0.012, 0.036]); this paired comparison is a descriptive companion rather than the preregistered primary. The mean-adjusted results in Table~\ref{tab:anchor_latent_dispersion} show that differences in average disclosure rates do not explain the lower variability under anchoring.
%Table~\ref{tab:anchor_latent_dispersion} demonstrates that accounting for the different average disclosure rates does not explain the reduction in variability under anchoring.

\subsection{Exact reference term and token scope}
The reference model is a frozen copy of the SFT checkpoint from which each RL run starts. For each sampled completion token, we use the non-negative k3 estimator with $r_{i,t}=\log\pi_{\mathrm{ref}}-\log\pi_\theta$. The added part of the accumulated loss is
\begin{equation}
\mathcal{L}_{\mathrm{anchor}}
=\frac{\beta}{32\cdot1024}
\sum_{i,t}m_{i,t}s_{i,t}\left[\exp(r_{i,t})-r_{i,t}-1\right],
\end{equation}
where $m_{i,t}$ marks active completion tokens. For failure-conditional anchoring, $s_{i,t}=1$ for every active token of a verifier-incorrect, format-valid rollout and is zero for solved, format-invalid, or empty rollouts. For all-token anchoring, $s_{i,t}=1$ for every active completion token. The denominator is always $32\cdot1024$, not the number of anchored tokens. The two arms thus use the same $\beta$ and per-token coefficient on different token sets; they do not have the same total regularization budget. At 1.5B, failure-conditional anchoring covers 0.570--0.601 of active completion tokens across runs, compared with 1.000 for all-token anchoring. The corresponding 7B range is 0.513--0.578. We did not adjust $\beta$ to match the total coefficient mass across arms.

\subsection{Progress-matched comparison}
Table~\ref{tab:progress-matched-anchor} compares the failure-conditional arm at step 300 with the unanchored checkpoint whose mean solve is closest to it. This is a post-hoc descriptive check, not a randomized control for every way an intervention might slow learning.

\begin{table*}[!tbp]
\caption{Progress-matched anchoring check for 1.5B Countdown. Slower capability progress explains part of the same-step narrowing, but disclosure remains less variable under anchoring than at the matched unanchored checkpoint.}
\label{tab:progress-matched-anchor}
\centering\small
\begin{tabularx}{\textwidth}{@{}Y c c Y@{}}
\toprule
Condition & Checkpoint & Disclosure SD & Role in comparison \\
\midrule
Unanchored & 300 & 0.189 & Same training horizon as anchoring \\
\rowrule
Unanchored & 150 & 0.139 & Closest mean solve to anchored step 300 \\
\rowrule
Failure-conditional anchor & 300 & 0.089 & Intervention result \\
\bottomrule
\end{tabularx}

\end{table*}

Relative to unanchored step 300, moving back to the progress-matched step-150 checkpoint reduces disclosure SD from 0.189 to 0.139. Anchoring reduces it further to 0.089, about 36\% below the progress-matched value. This comparison shows that slower progress accounts for part, but not all, of the observed narrowing; it does not fully separate the two effects.

\subsection{Coefficient ladder}
\begin{figure}[!htbp]
\centering
\includegraphics[width=0.50\columnwidth]{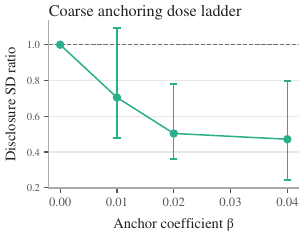}
\caption{Small reference-anchor coefficient sweep. Narrowing is established at $\beta=0.02$ but not at $\beta=0.01$; we do not treat this as a fitted dose-response law.}
\label{fig:anchor-dose}
\end{figure}

Figure~\ref{fig:anchor-dose} shows the small $\beta$ coefficient sweep used only as a sensitivity check, not as a fitted dose-response relationship.

\subsection{7B extension, provenance, and influence sensitivity}
\label{app:anchor-loo}
%The original 11 pairs were prospectively fixed on common-path 7B seeds. After their readout, we prospectively froze the remaining eight eligible seeds as an extension under the same $\beta=0.04$ intervention, mask, horizon, detector, probe, exclusions, and statistics. The eight-seed extension independently tightens disclosure dispersion, and the main text reports the pooled 19-pair estimate using a bootstrap stratified by experimental wave. Table~\ref{tab:anchor-7b-extension} reports the two waves separately and together.
We first selected 11 matched 7B model pairs before examining their outcomes. After analyzing those runs, we separately fixed the remaining eight eligible seeds as an extension, using exactly the same $\beta=0.04$ intervention, mask, training horizon, detector, probe, exclusion rules, and statistical analysis. The eight new pairs independently show reduced variation in disclosure across runs. The main text also reports the combined estimate across all 19 pairs, using a bootstrap that keeps the original and extension waves separate when resampling. Table~\ref{tab:anchor-7b-extension} reports the original wave, the extension, and the combined analysis. The original 11 pairs were directionally stable to every leave-one-run-out deletion, although interval-based significance was sensitive to four of the eleven runs; in the pooled 19 pairs the interval excludes zero after every deletion (Table~\ref{tab:anchor-loo}).

All 19 pairings pass the provenance check: no seed appears twice, each anchored run is paired with its pinned same-seed KL-off baseline, and all runs use the common execution configuration. The original 11 pairs used the common H200 execution configuration used for all 7B runs. The extension ran on a 4$\times$H200 pod after three replay checks reproduced the original checkpoint hashes bit-for-bit. A twelfth seed in the original program changed batch geometry after an out-of-memory error and was excluded before any outcome was seen; no licensed run was removed after readout.

\begin{table*}[!tbp]
\caption{7B anchoring extension and pooled result. The extension was frozen after the original readout and reproduces the tightening itself. The pooled interval uses a wave-stratified bootstrap.}
\label{tab:anchor-7b-extension}
\centering\small
\begin{tabularx}{\textwidth}{@{}Y c c c c@{}}
\toprule
Run set & n & Log SD ratio [97.5\% interval] & SD: anchor / no anchor & Change in solve \\
\midrule
Original fixed pairs & 11 & $-1.36$ [$-3.13$, $-0.36$] & 0.062 / 0.242 & +0.062 [$-0.021$, +0.145] \\
\rowrule
Frozen extension & 8 & $-1.01$ [$-1.75$, $-0.32$] & 0.111 / 0.305 & +0.080 [$-0.020$, +0.179] \\
\rowrule
Pooled, wave-stratified & 19 & $-1.14$ [$-1.55$, $-0.84$] & 0.090 / 0.281 & +0.069 [+0.012, +0.126] \\
\bottomrule
\end{tabularx}

\end{table*}

\begin{table*}[!tbp]
\caption{Leave-one-run-out sensitivity for the 7B anchoring result. The original and extension-only results are each based on few pairs, while the pooled $n=19$ estimate excludes zero after every one-run deletion.}
\label{tab:anchor-loo}
\centering\small
\begin{tabularx}{\textwidth}{@{}Y c c Y@{}}
\toprule
Analysis & n & Log SD ratio or deletion range & Interval stability \\
\midrule
Original pairs & 11 & $-1.36$ [$-3.13$, $-0.36$] & Four of 11 deletion intervals cover zero. \\
\rowrule
Extension alone & 8 & $-1.01$ [$-1.75$, $-0.32$] & Four of eight deletion intervals cover zero. \\
\rowrule
Pooled pairs & 19 & $-1.14$ [$-1.55$, $-0.84$] & Full interval excludes zero. \\
\rowrule
Pooled leave-one-out & 19 deletions & Point estimates $-1.26$ to $-1.06$ & All 19 deletion intervals exclude zero. \\
\bottomrule
\end{tabularx}

\end{table*}

%\section{Mid-RL exact-state anchoring and localization}
\section{Anchoring, later-RL divergence, and localization}
\label{app:kbr}

\subsection{Anchored-history continuation and post-history anchoring}
\begin{figure*}[!tbp]
\centering
\includegraphics[width=0.86\textwidth]{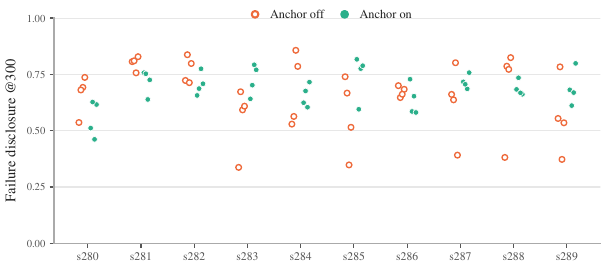}
\caption{All KBR continuation runs across ten mid-RL parent states with anchored histories. Each parent has four futures with anchoring switched off and four with anchoring kept on.}
\label{fig:app-kbr-full}
\end{figure*}

\begin{figure}[!htbp]
\centering
\includegraphics[width=0.62\columnwidth]{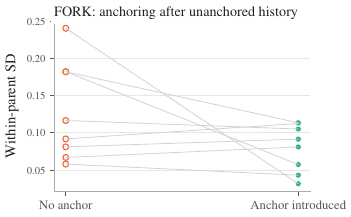}
\caption{FORK variability after an unanchored training history. Starting from each of eight exact mid-RL parent states, we run four matched future continuations without anchoring and four with anchoring introduced at the fork. Each dot is the within-parent SD of failure disclosure across those four continuations; lower values therefore indicate less branching across otherwise matched futures.}
\label{fig:app-fork}
\end{figure}

Table~\ref{tab:kbr-fork} compares two mid-RL continuation designs: the anchored-history continuation test (KBR) continues or switches off anchoring after an anchored history, while the post-history anchoring test (FORK) introduces anchoring after an unanchored history. Figure~\ref{fig:app-kbr-full} shows all KBR continuation children, and Figure~\ref{fig:app-fork} shows the smaller FORK summary across parent mid-RL states.

\begin{table*}[!tbp]
\caption{Mid-RL continuation tests of anchoring. KBR shows reduced later-RL divergence after an anchored history. FORK does not establish narrowing on the raw scale when anchoring is introduced only after an unanchored history.}
\label{tab:kbr-fork}
\centering\small
\begin{tabularx}{\textwidth}{@{}Y c c c Y@{}}
\toprule
Design & Parents & SD ratio & log SD ratio interval & Interpretation \\
\midrule
Anchored-history continuation (KBR): keep anchor on vs switch off & 10 & 0.47 & $-0.75$ [$-1.22$, $-0.25$] & 8/10 parents show lower variability with anchoring on \\
\rowrule
Post-history anchoring (FORK): introduce anchor after unanchored history & 8 & 0.60 & $-0.51$ [$-1.10$, +0.09] & Reduced variability is not statistically established \\
\bottomrule
\end{tabularx}

\end{table*}

For KBR, continuing anchoring reduces pooled within-parent disclosure SD from 0.138 to 0.065 (SD ratio 0.47; log ratio $-0.75$ [$-1.22$, $-0.25$]), with lower variability in 8/10 parent states. Mean disclosure is not detectably changed (0.683 with anchoring on versus 0.658 with anchoring off), while probe solve shows a detected 4-point decrease, from 0.386 to 0.344 ($\Delta=-0.043$). This conclusion is history-specific: the complementary FORK experiment, which introduces anchoring only after an unanchored history, does not establish reduced variability.

\subsection{Token-region localization}
Table~\ref{tab:localization} compares full anchoring with anchoring only the tail, the reporting seam, or the failed-reasoning portion. Tables~\ref{tab:locklan-gradient} and~\ref{tab:locklan-drift} report the corresponding gradient and reference-drift analyses.

\begin{table*}[!tbp]
\caption{Where the reference constraint acts. Anchoring only failed reasoning recovers the full stabilization effect and capability cost. Short tail and seam masks do not establish narrowing and cover far fewer tokens. Intervals are 97.5\% for the full anchor and 98.33\% (a three-arm family) for the failed-reasoning, tail, and seam rows.}
\label{tab:localization}
\centering\small
\begin{tabularx}{\textwidth}{@{}Y c c c@{}}
\toprule
Anchor region & log SD ratio vs sham & Change in solve & Active-token exposure \\
\midrule
Full failure-conditional & $-0.7514$ [$-1.4200$, $-0.2250$] & $-0.0597$ [$-0.0888$, $-0.0306$] & 58.9\% \\
\rowrule
Failed reasoning & $-0.7109$ [$-1.2055$, $-0.1834$] & $-0.0643$ [$-0.0859$, $-0.0427$] & 53.6\% \\
\rowrule
Post-attempt tail & $-0.0468$ [$-0.6499$, +0.6019] & $-0.0331$ [$-0.0731$, +0.0069] & 4.3\% \\
\rowrule
16-token seam & $-0.0438$ [$-0.5230$, +0.5503] & $-0.0038$ [$-0.0345$, +0.0270] & 1.7\% \\
\bottomrule
\end{tabularx}
\end{table*}

\begin{table*}[!tbp]
\caption{Gradient and drift localization analysis (LOCKLAN). The failed-reasoning-only gradient is almost as large as the full-anchor gradient and points in nearly the same direction, while tail and seam gradients are much smaller.}
\label{tab:locklan-gradient}
\centering\small
\begin{tabularx}{\textwidth}{@{}Y c c c c@{}}
\toprule
States & $\lVert g_D\rVert / \lVert g_{\mathrm{full}}\rVert$ & $\cos(g_D,g_{\mathrm{full}})$ & Tail norm ratio & Seam norm ratio \\
\midrule
Anchored @ 150 & 0.960 & 0.983 & 0.173 & 0.132 \\
Anchored @ 300 & 0.964 & 0.984 & 0.172 & 0.137 \\
Sham @ 150 & 0.975 & 0.993 & 0.110 & 0.095 \\
Sham @ 300 & 0.958 & 0.989 & 0.139 & 0.103 \\
\bottomrule
\end{tabularx}
\end{table*}

\begin{table*}[!tbp]
\caption{Teacher-forced drift from the reference policy at step 300. About 92--95\% of the measured drift lies before the reporting seam in the reported 1.5B and 7B settings.}
\label{tab:locklan-drift}
\centering\small
\begin{tabular}{@{}l c c c c@{}}
\toprule
Setting @300 & Pre-seam drift mass & Seam mass & Other tail mass & Total drift \\
\midrule
1.5B / KL-off & 0.934 & 0.037 & 0.029 & 0.119 \\
1.5B / Sham & 0.934 & 0.038 & 0.028 & 0.119 \\
1.5B / Anchor & 0.947 & 0.029 & 0.024 & 0.031 \\
7B / outcome-only & 0.923 & 0.058 & 0.019 & 0.199 \\
7B / Anchor & 0.953 & 0.025 & 0.023 & 0.047 \\
\bottomrule
\end{tabular}
\end{table*}

\FloatBarrier
%\clearpage
\section{Extended robustness and boundary analyses}
\label{app:robustness}
Table~\ref{tab:reviewer-concerns} links common alternative explanations and scope questions to the corresponding evidence elsewhere in the appendix.

\begin{table}[H]
\caption{Summary of robustness and scope evidence.}
\label{tab:reviewer-concerns}
\centering\small
\begin{tabularx}{\textwidth}{@{}Y Y Y@{}}
\toprule
Question & Evidence & Main takeaway \\
\midrule
Role of reference KL & KL-free setup + all-token and failure-conditional anchoring & The phenomenon is measured in a deliberate KL-free regime; restoring reference regularization narrows both readouts, especially disclosure. \\
\rowrule
Generic RL variability & Primary + OLMo + shortest-path selective-dispersion comparisons & Variability depends on the behavior: disclosure is much less reproducible than solve in the clean experiments. \\
\rowrule
Capability differences & Clean 1.5B/OLMo experiments; 7B boundary & Capability is tightly reproduced in the clean experiments; 7B is explicitly treated as a broader persistence boundary. \\
\rowrule
Detector sensitivity & Human audits + detector-tier sensitivity & Human agreement is high, randomized audit frames are included, and broader detectors preserve the qualitative spread. \\
\rowrule
Probe selection & Full dev-100 disclosure + selected-20 versus other-80 split & Broad spread remains on the 80 prompts outside the behavior-selected probe; selection does not explain the result. \\
\rowrule
Fixed-horizon interpretation & Nine-rung trajectory + small longer-horizon cases & The claim is reproducibility at a fixed budget, not separate stable endpoints; longer outcomes are mixed (\ref{app:subsec:long_horizon}). \\
\rowrule
Report-entry localization & NoBreak control + mean-adjusted Step B + route-scale check & Most tightening comes from the learned paragraph-break cue; this is template-specific and the 70\% route split is raw-scale only. \\
\rowrule
Dependence on report SFT & Instruction-conditioned Qwen-32B runs & Broad reporting variation also appears without the primary report-SFT setup. \\
\rowrule
Algorithm scope & Stabilized PPO n=8; RLOO descriptive only & PPO provides lower-variance boundary evidence; we do not claim algorithm invariance from RLOO. \\
\rowrule
Task and model scope & Shortest-path + OLMo-1B & The phenomenon is not limited to Countdown or Qwen. \\
\rowrule
Anchoring level effects & Paired level effects + 7B retention & Anchoring narrows variation without lowering the mean in the primary runs; at 7B it mainly preserves the reference reporting behavior. \\
\rowrule
Mid-RL state restoration & Same-key twins + same-configuration replay & Matched restores and replays are bit-exact in the replay checks. \\
\bottomrule
\end{tabularx}

\end{table}

For the primary runs, robust dispersion measures preserve the ordering (MAD ratio 2.87; IQR ratio 3.02), and fixed failure-mixture reweighting changes the disclosure span only from 0.646 to 0.629. For the shortest-path task, the comparison on the transformed scale is treated as inconclusive because probe-solving performance is already close to its floor; we therefore do not generalize from it to the cleaner experiments. Robust measures of variability and leave-one-out checks are discussed in the main text only when they meaningfully qualify a main claim.%For shortest-path, the transformed-scale selective comparison is treated as a non-detection on the floor-adjacent probe solve surface rather than generalized from the cleaner experiments. Robust-dispersion and leave-one-out checks are cited in the main text only when they meaningfully qualify a main claim.

\subsection{Training beyond 300 steps}
\label{app:subsec:long_horizon}
%Our primary estimates describe a fixed step-300 horizon rather than an asymptotic endpoint. Table~\ref{tab:long-horizon} summarizes the multi-run cases trained beyond step 300. No row contains more than five same-instrument runs, and one row pools two recipes only for a descriptive comparison, so these cases do not establish a population trend.
Our main results are measured at step 300, so they show what happens at that point in training rather than what would necessarily happen after much longer training. Table~\ref{tab:long-horizon} summarizes the few cases that were continued beyond step 300. Each case includes at most five runs measured in the same way, and one row combines two training recipes only for a descriptive comparison, so these results are too limited to support a general long-term trend.

\begin{table*}[!tbp]
\caption{Small multi-run cases trained beyond step 300. SD is calculated across runs using the detector named in each row. These values are descriptive because every case has $n\leq5$ and the instruments differ across model families.}
\label{tab:long-horizon}
\centering\small
\begin{tabularx}{\textwidth}{@{}Y c c c c Y@{}}
\toprule
Setting & $n$ & Last step & Disclosure SD, 300 $\rightarrow$ last & Solve SD, 300 $\rightarrow$ last & Last-step disclosure \\
\midrule
Qwen2.5-1.5B, outcome only & 3 & 600 & 0.273 $\rightarrow$ 0.023 ($D_{v1}$) & 0.064 $\rightarrow$ 0.131 & 0.039 / 0.000 / 0.000 \\
\rowrule
Qwen2.5-1.5B, length penalty & 3$^{*}$ & 600 & 0.092 $\rightarrow$ 0.148 ($D_{v2}$) & 0.050 $\rightarrow$ 0.217 & 0.404 / 0.651 / 0.667 \\
\rowrule
Qwen2.5-7B, outcome only & 5 & 600 & 0.306 $\rightarrow$ 0.396 ($D_{v1}$) & 0.075 $\rightarrow$ 0.201 & 0.000--0.918 \\
\rowrule
OLMo-2-7B, outcome only & 3 & 1200 & 0.099 $\rightarrow$ 0.169 $\rightarrow$ 0.442 ($D_{v5}$) & 0.081 $\rightarrow$ 0.163 $\rightarrow$ 0.030 & 0.873 / 0.000 / 0.555 \\
\rowrule
Gemma-2-9B, outcome only & 3 & 600 & 0.221 $\rightarrow$ 0.115 ($D_{v5}$) & 0.105 $\rightarrow$ 0.244 & 0.666 / 0.886 / 0.833$^{\dagger}$ \\
\rowrule
Distilled Qwen-1.5B, two recipes & 2+2$^{\ddagger}$ & 600 & 0.049 $\rightarrow$ 0.078 ($D_{v2}$) & 0.055 $\rightarrow$ 0.013 & 0.797--0.961 \\
\midrule
\multicolumn{6}{@{}p{0.98\textwidth}@{}}{\footnotesize $^{*}$Three runs spanning two seeds and two execution configurations. $^{\dagger}$The apparent narrowing is detector-dependent: $D_{v1}$ gives 0.676 / 0.004 / 0.787 at step 600. $^{\ddagger}$Two outcome-only and two length-penalty runs are pooled only for this descriptive row, not as one combined estimate.} \\
\bottomrule
\end{tabularx}

\end{table*}

The later outcomes are mixed. Three outcome-only 1.5B Qwen runs move toward very low primary-detector disclosure by step 600 (0.039, 0.000, and 0.000), although the broadest detector still reads 0.103, 0.161, and 0.215 and individual runs pass through both high- and low-disclosure states along the way. Five outcome-only 7B runs still span 0.000--0.918 at step 600. Three OLMo-2-7B runs have disclosure SD 0.169 at step 600, then end at 0.873, 0.000, and 0.555 at step 1200. A capability change in one run makes the later comparison harder to interpret.

Detector choice also matters at long horizons. In one distilled 1.5B run continued to step 1200, the narrow phrase detector falls to 0.067 while broader content detectors remain at 0.882--0.914. Thus a surface-form change can look like a loss of disclosure under a narrow lexical monitor. Overall, these small cases show no single post-300 endpoint: reporting can erode, remain mixed, or change wording. Our primary claim hence concerns reproducibility at a fixed training budget, not distinct stationary endpoints. A possible explanation is that disclosure erodes at different rates across runs as training continues. The three 1.5B cases are compatible with this, but too few to establish it. %One possibility is that anchoring's effect weakens at different rates across runs. The three 1.5B control cases are compatible with that explanation, but they are too limited to either confirm or rule it out.%A heterogeneous-erosion account is consistent with the three 1.5B controls and is neither established nor excluded by them.

\clearpage

\section{Extended related work}
\label{app:related-extended}

\paragraph{Failure awareness, self-correction, and retraction.}
A large body of work studies whether models can recognize errors or improve after noticing them. \citet{kadavath2022know} study whether models can estimate if they know an answer; \citet{tyen2024errors} separate finding an error from correcting it; and recent work studies self-verification, retraction, correction, and stepwise failure detection under different forms of feedback \citep{yang2026admit,chen2026selfverify,tsui2026selfcorrectionbench,huang2024selfcorrect,kamoi2024survey,mavi2025selfevaluating}. Our question is not whether the model can eventually fix the answer, but whether an objectively failed attempt is reported consistently across retrainings.

\paragraph{Honesty, abstention, and reporting incentives.}
\citet{joglekar2025confessions} train a separate confession channel; \citet{zhai2026abstainr1} optimize abstention under RLVR; and \citet{kalai2026hallucinations} study incentives that can reward guessing instead of admitting uncertainty. \citet{wen2025mislead} report that RLHF can make wrong answers more persuasive, while \citet{chandna2026misleadvalid} highlights validity concerns in the original reward-model setup. Work on verbalized overconfidence and confidence--correctness alignment provides additional context \citep{zhao2026overconfidence,xie2026knowwrong}. %Our setting instead asks whether an existing, unrewarded reporting behavior remains reproducible through post-training.
Our primary setting instead asks whether an existing reporting behavior that is not rewarded by the task objective remains reproducible through post-training; auxiliary experiments then compare direct reward with reference-based preservation.

\paragraph{Underspecification and numerical reproducibility.}
Our broad framing follows work on underspecification: standard evaluation metrics can leave other behaviors poorly constrained \citep{damour2022underspecification}. Sensitivity to random seeds is also well documented in RL and language-model training \citep{henderson2018deeprl,fehlauer2025seeds,bui2025randomseeds}. Formal work studies replicable RL \citep{eaton2026replicable}, while systems work shows that floating-point ordering, kernels, and implementation details can introduce nondeterminism \citep{yuan2025nondeterminism,shanmugavelu2024floatingpoint,gond2026llm42,he2025nondeterminism,whitehead2011floatingpoint}. \citet{altintas2025butterfly} provide a close comparison to our trajectory framing, showing that tiny perturbations can grow during training. We ask whether one safety-relevant behavior is selectively less reproducible than task capability and whether reporting can still diverge later in RL from the same training history.

\paragraph{Post-training behavioral drift and preservation.}
Recent work studies safety-relevant behaviors that can emerge or persist during post-training \citep{schreiber2026overtrained,betley2025emergent,betley2026narrowtasks}. Reference regularization can limit behavioral drift: \citet{elcock2026taskadaptation} study preservation during task adaptation, while \citet{soligo2026narrow} examine how KL constraints can favor a narrower set of solutions. It is not automatically benign for oversight: \citet{golechha2026klpenalties} report that KL penalties can increase chain-of-thought unfaithfulness in reward-hackable coding environments, motivating our treatment of anchoring as a tradeoff rather than a free constraint. Closer to our reproducibility question, \citet{hussing2026behaviorconsistent} formalize behavioral consistency across deep-RL runs and reduce policy differences using adaptive entropy regularization toward a common prior. We instead apply a reference term to a measured reproducibility failure in language-model RL, test it across retrainings and controlled mid-RL continuations, and study which part of failed trajectories carries the stabilizing effect.

\paragraph{Credit assignment, verifiers, and monitorability.}
Process-supervision methods such as rewarding verified prefixes motivate our local credit experiments \citep{liu2026goodprefix}. Work on gaming verifiers also shows that even a formally checkable objective can leave intended behavior underspecified \citep{helff2026gamingverifiers}. Chain-of-thought monitorability research asks which optimization pressures preserve useful behavioral traces \citep{macdermott2025pressure,guan2026monitorability,emmons2025necessary}, while related work cautions that textual ``verification'' can be causally weak \citep{zhao2026aha}. These works motivate preserving reporting behavior, but we do not equate failure disclosure with faithful reasoning or grounded verification.

\paragraph{Hallucination-adjacent framing.} Failure disclosure is related to, but distinct from, hallucination or confident fabrication. Recent definitions emphasize observable mismatch with a relevant world model or reference \citep{liu2026hallucinationdefinition,liu2026halluworld}. We instead ask what the model says about its own unsuccessful attempt in a setting where task failure is objectively known. We therefore avoid calling nondisclosure ``lying'' or treating failure disclosure as simply a type of hallucination.

\end{document}